\documentclass[smallextended]{svjour3}       
\smartqed  
\usepackage[]{amssymb,mathtools}
\usepackage{microtype}
\usepackage{graphicx}
\usepackage{subfigure}
\usepackage{booktabs} 
\usepackage{algorithm}
\usepackage{nomencl}
\makenomenclature
\usepackage{algpseudocode}
\usepackage{wrapfig}
\allowdisplaybreaks
\usepackage[utf8]{inputenc}
\DeclareUnicodeCharacter{2212}{-}
\begin{document}

\title{High-dimensional Bayesian optimization using low-dimensional feature spaces
}

\titlerunning{High-dimensional Bayesian optimization using low-dimensional feature spaces}        

\author{Riccardo Moriconi                
\and
        Marc~Peter~Deisenroth  
        \and
        K.~S.~Sesh~Kumar 
}


\institute{Riccardo Moriconi \at
              Department of Computing \\
              Imperial College London, UK\\
                         \email{r.moriconi16@imperial.ac.uk} 
           \and
           \mbox{Marc Peter Deisenroth} \at
              Department of Computer Science \\
              University College London, UK \\
                         \email{m.deisenroth@ucl.ac.uk}
              \and
           K.~S.~Sesh~Kumar \at
              Data Science Institute \\
              Imperial College London, UK\\
            \email{s.karri@imperial.ac.uk}
}

\date{}

\maketitle

\begin{abstract}
Bayesian optimization (BO) is a powerful approach for seeking the global optimum of expensive black-box functions and has proven successful for fine tuning hyper-parameters of machine learning models. However, BO is practically limited to optimizing 10--20 parameters. To scale BO to high dimensions, we usually make structural assumptions on the decomposition of the objective and\slash or exploit the intrinsic lower dimensionality of the problem, e.g. by using linear projections. We could achieve a higher compression rate with nonlinear projections, but learning these nonlinear embeddings typically requires much data. This contradicts the BO objective of a relatively small evaluation budget. To address this challenge, we propose to learn a low-dimensional feature space jointly with (a) the response surface and (b) a reconstruction mapping. Our approach allows for optimization of BO's acquisition function in the lower-dimensional subspace, which significantly simplifies the optimization problem. We reconstruct the original parameter space from the lower-dimensional subspace for evaluating the black-box function. For meaningful exploration, we solve a constrained optimization problem\footnote{Implementation available at https://github.com/rm4216/BayesOpt}.
\end{abstract}

\section{Introduction}
Bayesian optimization (BO) is a useful model-based approach to global optimization of black-box functions, which are expensive to evaluate \cite{Kushner1964,JonesBO1998}. This sample-efficient technique for optimization has been effective in experimental design of machine learning algorithms~\cite{Bergstra2011}, robotics applications~\cite{Cully2015,CalandraGait2016} and medical therapies~\cite{Sui2015} for optimization of spinal-cord electro-stimulation. Despite its great success, BO is practically limited to optimizing 10--20 parameters. A large body of literature has been devoted to address scalability issues to elevate BO to high-dimensional optimization problems, such as discovery of chemical compounds \cite{GomezBombarelliAutomatic2017} or automatic software configuration \cite{Hutter2011b}.

The standard BO routine consists of two key steps: (i) estimating the black-box function from data through a probabilistic surrogate model, usually a Gaussian process~(GP), referred to as the \emph{response surface}; (ii) maximizing an \emph{acquisition function} that trades off exploration and exploitation according to uncertainty and optimality of the response surface. 
As the dimensionality of the input space increases, these two steps become challenging. The sample complexity to ensure good coverage of inputs for learning the response surface is exponential in the number of dimensions \cite{bo_review16}. With only a small evaluation budget, the learned response surface and the resulting acquisition function are characterized by vast flat regions interspersed with highly non-convex landscapes \cite{ElasticBO}. 
This renders the maximization of the acquisition in high dimensions inherently hard \cite{Garnett2013ActiveProcesses}.

High-dimensional optimization is often translated into low-dimensional problems, which are defined on subsets of variables~\cite{moriconi2019high,kadasamy_add,rolland2018high}. These approaches apply a divide and conquer approach to decompose the problem into independent~\cite{moriconi2019high,kadasamy_add} and potentially dependent components~\cite{rolland2018high}. 
However, high-dimensional data often possesses a lower intrinsic dimensionality, which can be exploited for optimization.
A feature mapping can then be used to map the original $D$-dimensional data onto a $d\ll D$-dimensional manifold. For example, in~\cite{Wang2013b}, the authors used random linear mappings to reduce dimensionality of the optimization problem. Similar approaches, which use linear dimensionality reduction, drive exploration in BO to actively learn this linear embedding \cite{Garnett2013ActiveProcesses}. While these methods perform well in practice, they are restricted to linear subspaces of the original domain. With nonlinear embeddings, higher compression rates are possible. In our work, we focus on this nonlinear setting.

Using BO with nonlinear feature spaces was proposed in \cite{GomezBombarelliAutomatic2017,gonzalez2015bayesian,kusner2017grammar,LobatoChemical}. In \cite{GomezBombarelliAutomatic2017}, a low-dimensional data representation is learned with variational autoencoders (VAEs) \cite{JimenezRezende2014a,Kingma2014a}. 
However, this approach requires both large amounts of data and learning the model offline without the possibility to update the learnt feature space during optimization. Nevertheless, in the specific application of automatic discovery of molecules, where libraries of existing compounds are readily available prior to optimization, this approach makes much sense. To accommodate fairly small evaluation budgets, in our work, we exploit a  probabilistic model based on GPs, which features superior data efficiency with respect to VAE-based approaches~\cite{GomezBombarelliAutomatic2017,gonzalez2015bayesian,kusner2017grammar,LobatoChemical}.
VAE models \cite{lu2018structured} were used to propagate uncertainty of latent space representations through the response surface model with Gaussian process latent variable models~\cite{lawrence2005probabilistic,Titsias2010,Lawrence2006}. However, in \cite{lu2018structured}, the latent space representation is not learned specifically for the regression task (learning the response surface). Gradient-based methods~\cite{abbati2018adageo} have been used to learn a lower-dimensional Riemannian manifold for optimization and sampling.

Nonlinear embeddings also allow for modeling non-stationary objective functions. In this context, a hierarchical composition of GPs, referred to as \emph{deep GPs}~\cite{deep0damianou2013,deep1salimbeni2017doubly,deep2dai2015variational,deep3damianou2015deep,deep4hensman2014nested}, is especially useful when the response surface is characterized by abrupt changes or has constraints. An extensive investigation on the employment of deep GP models in BO is presented in \cite{deep2dai2015variational,hebbal2019bayesian}. In our work, we also exploit the idea of learning highly nonlinear functions through the composition of simpler functions \cite{lecun2015deep}, but we focus on deterministic dimensionality reduction and optimization in feature space.

\begin{figure}
    \centering
    \includegraphics[width=0.4\hsize]{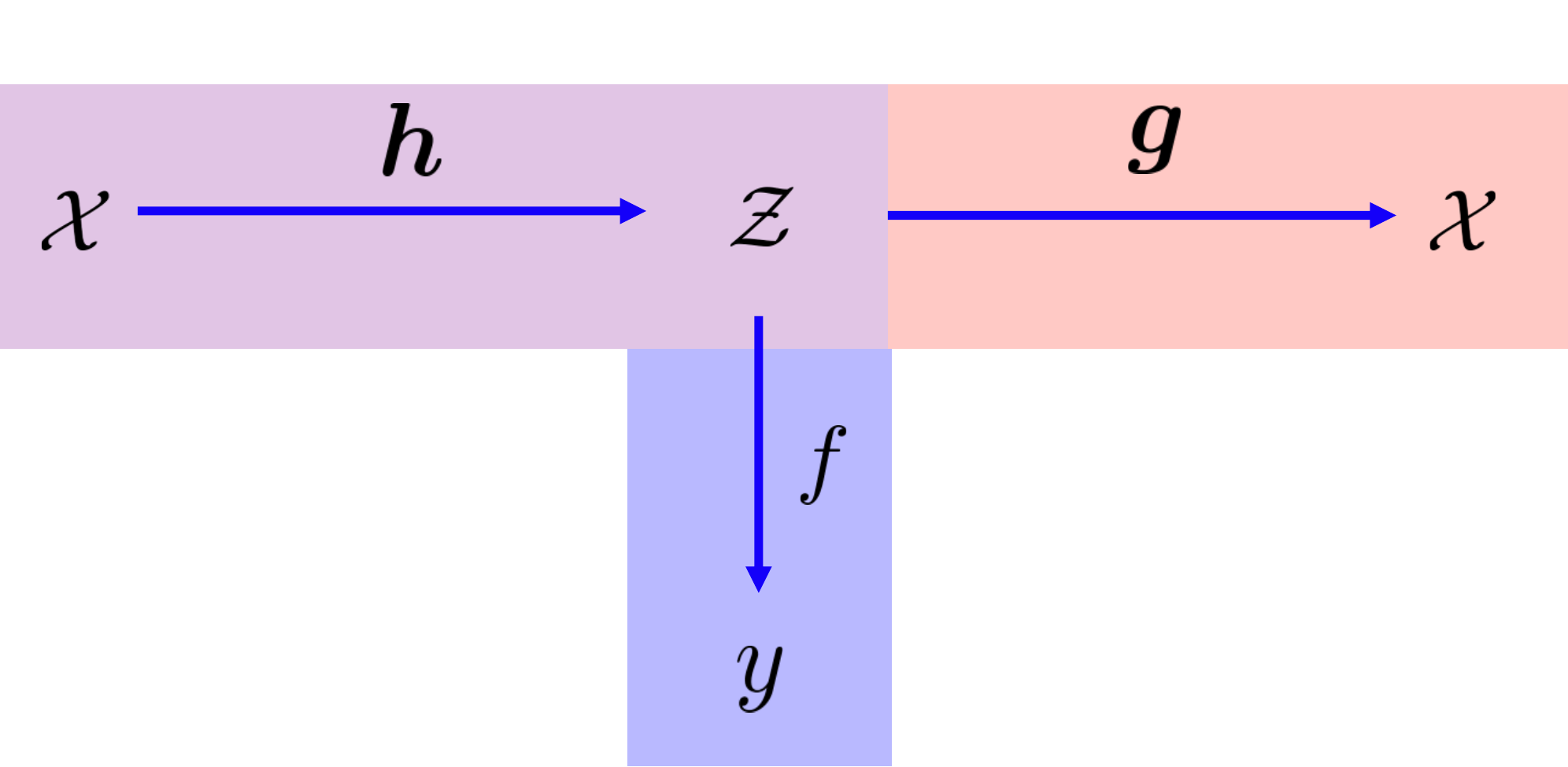}
    \caption{Model for Bayesian optimization on data manifolds, jointly solving two distinct tasks: (i) a regression from feature space to observations (in blue) and (ii) a reconstruction mapping from feature space to high-dimensional space (in red).}
        \label{fig:GP-AE}
\end{figure}
In this paper, we propose a BO algorithm for high-dimensional optimization, which learns a nonlinear feature mapping $\mathbf{h}:\mathbb{R}^D\to\mathbb{R}^d$ to reduce the dimensionality of the inputs, and a \emph{reconstruction mapping} $\mathbf{g}\colon\mathbb{R}^{d}\rightarrow\mathbb{R}^{D}$ based on GPs to evaluate the true objective function, jointly, see Figure~\ref{fig:GP-AE}. This allows us to optimize the acquisition function in a lower-dimensional feature space, so that the overall BO routine scales to high-dimensional problems that possess an intrinsic lower dimensionality. Finally, we use constrained maximization of the acquisition function in feature space to prevent meaningless reconstructions.

\section{Bayesian optimization}

\begin{algorithm}
\caption{Key steps of Bayesian optimization in feature space. The response surface learning and acquisition function maximization are performed in feature space with dimensionality $d\ll D$. The reconstruction step in line \ref{reconstruct} allows us to run experiments with the original objective function, $f_{X}$.}
\label{FeatureBO}
  \begin{algorithmic}[1]
   	\State {\bfseries Inputs:} $\mathbf{X}_{0}\in\mathbb{R}^{N_{0}\times D}$, $\mathbf{y}_{0}\in\mathbb{R}^{N_{0}}$
      \For{$t = 0,1,2,...,T_{\text{end}}$}
        \State \textbf{Response surface learning}
        \State $\ \ \ f_X = f_Z\circ \mathbf{h}$\label{macro_i} 
        \Comment{Composition of feature map and low-dimensional response surface}
        \State $\ \ $ $\mathbf{Z}_{t}=\mathbf{h}(\mathbf{X}_{t})$\label{h_map}
        \Comment{Dimensionality reduction}
        \State $\ \ $ $p(f_{Z}|\mathbf{Z}_{t}, \mathbf{y}_{t})$\label{f_Zmap}
        \Comment{Learning of the response surface in low-dimensional feature space}
        \State \textbf{Optimal input selection }$\mathbf{x}_{t+1}$\label{macro_ii}
        \State $\ \ $ $\mathbf{z}_{*} = \underset{\mathbf{z}\in\mathcal{Z}}{\text{argmax}}\  \alpha(\mathbf{z})$\label{acq_max}
        \Comment{Acquisition function maximization in feature space} 
        \State $\ \ $ $\mathbf{x}_{t+1} := \mathbf{g}(\mathbf{z}_{*})$\label{reconstruct}
        \Comment{Reconstruction of high-dimensional input}
        \State \textbf{Evaluation}\label{macro_iii}
        \State $\ \ $ $y_{t+1} = f_{X}(\mathbf{x}_{t+1})+\varepsilon$
        \Comment{Evaluation of noisy high-dimensional objective function}
        \State $\mathbf{X}_{t}\cup\{\mathbf{x}_{t+1}\}$, \quad $\mathbf{y}_{t}\cup\{y_{t+1}\}$
      \EndFor
      \State \textbf{Return} $\mathbf x^{*} = \arg\min \mathbf{y}_{t}$
      \Comment{Minimizer of the objective function $f_{X}$}
  \end{algorithmic}
\end{algorithm}

Bayesian optimization is a powerful tool for globally optimizing black-box functions that are expensive to evaluate \cite{JonesBO1998,Kushner1964,Mockus1975OnExtremum}. In our setting, we consider the global minimization problem
\begin{equation}
    \mathbf{x}^{*} = \underset{\mathbf{x}\in\mathcal{X}}{\arg\min}\, f_{X}(\mathbf{x})
\end{equation}
with input space $\mathcal{X}=[0,1]^{D}$ and objective function $f_{X}\colon\mathcal{X}\rightarrow\mathbb{R}$. We consider functions $f_{X}$ that are costly to evaluate and for which we are allowed a small budget of evaluation queries to express our best guess of the optimum's location $\mathbf{x}^{*}$ in at most $T_{\text{end}}$ iterations. We further assume we have access only to noisy evaluations of the objective $y=f_{X}+\varepsilon$, where $\varepsilon\sim\mathcal{N}(0, \sigma_{n}^{2})$ is i.i.d. Gaussian measurement noise with variance $\sigma_n^2$. 
We restrict ourselves to the typical setting, where neither gradients nor convexity properties of $f_X$ are available.

 The main steps of a BO routine at iteration $t$ involve (i) \emph{response surface learning}, (ii) \emph{optimal input selection} $\vec x_{t+1}$ and (iii) \emph{evaluation} of the objective function $f_X$ at $\vec x_{t+1}$. The first step trains a probabilistic surrogate model $p(f_{X})$, the response surface, which describes the black-box relationship between inputs $\vec x$ and observations $y$. In the $(t+1)$st iteration of BO, the optimal input selection step finds an input $\vec x_{t+1}$ that maximizes an \emph{acquisition function} $\alpha(\cdot)$, which describes the added value of input $\vec x_{t+1}$. The evaluation step returns a noisy observation of the true objective function $f_{X}(\vec x_{t+1})+\varepsilon$ at the selected location. These steps are summarized in lines \ref{macro_i}, \ref{macro_ii} and \ref{macro_iii} of Algorithm~\ref{FeatureBO}, respectively. Having defined a probabilistic surrogate model for our objective function, which is usually modeled by a GP~\cite{Rasmussen2006}, we can compute posterior predictions of objective function values at test locations. These posterior predictions are then fed to the acquisition function, which drives exploration during optimization. Posterior predictions of the GP are Gaussian distributed with mean $\mu$ and variance $\sigma^2$. Defining $Z(\vec x):=(f_{\text{min}} - \mu(\vec x))/\sigma(\vec x)$ and $f_{\text{min}}:=\underset{\vec x \in \vec X_{t}}{\min} f(\vec x)$, this  allows us to define three different acquisition functions to maximize:
 \begin{align}
     \hspace{-3mm}\alpha(\vec x) &\!=\!\Phi(Z(\vec x)) \hspace{30mm} \text{ Probability of improvement (PI)~\cite{Kushner1964}}\label{equationPI}\\
   \hspace{-3mm}\alpha(\vec x) &\!=\!\sigma(\vec x)Z(\vec x)\Phi(Z(\vec x)) \!+\! \sigma(\vec x)\phi(Z(\vec x)) \text{ Expected improvement (EI)~\cite{Mockus1975OnExtremum}}  \label{equationEI}\\
    \hspace{-3mm}\alpha(\vec x) &\!=\!-\mu(\vec x)\! +\! \beta_{t}\sigma(\vec x)\hspace{20.5mm} \text{ Upper confidence bound (UCB)~\cite{Srinivas2010}} \label{equationUCB}.
\end{align}
Here, $\phi$ and $\Phi$ denote the probability density function and the cumulative density function of the standard normal $\mathcal N(0,1)$, respectively. The parameter $\beta_{t}$ controls the exploration exploitation trade-off. For a complete review on acquisition function the reader is referred to \cite{bo_review16}. In high-dimensional settings ($D>20$), both the response surface learning and optimal input selection via optimization of the acquisition function are computationally challenging.

\section{Bayesian optimization in low-dimensional feature spaces}
In this section, we consider a setting, where the input space $\mathcal{X}$ is high-dimensional and the objective function possesses an intrinsic lower dimensionality. In our work, we exploit the effective low dimensionality of the objective function for BO in a lower-dimensional \emph{feature space} $\mathcal{Z}\subset\mathbb{R}^{d}$, where $d\ll D$. In particular, we express the true objective $f_X:\mathbb{R}^D\to \mathbb{R}$ as a composition of a feature mapping $\mathbf{h}:\mathbb{R}^D\to\mathbb{R}^d$ and a  function $f_{Z}\colon\mathcal{Z}\rightarrow\mathbb{R}$ so that $f_X = f_Z\circ \vec h$. The lower-dimensional feature space allows for both learning the response surface $f_X$ and maximizing an acquisition function $\alpha$ with domain $\mathcal Z$, which yields optimizer $\vec z_*$. 
%
Since we cannot evaluate the true objective $f_X$ directly at the low-dimensional features $\vec z_*$, we project $\vec z_*$ back into the $D$-dimensional data space  $\mathcal{X}$ by means of a \emph{reconstruction} mapping $\mathbf{g}\colon\mathcal{Z}\rightarrow{\mathcal{X}}$. We can think of this mapping as a decoder within an auto-encoder framework. We model both the composition $f_X := f_Z\circ \mathbf{h}$ and the reconstruction with GPs \cite{Rasmussen2006}. Algorithm~\ref{FeatureBO} summarizes the main steps of this feature-space BO.

In the following, we detail the model (see Figure~\ref{fig:GP-AE}) for jointly learning the feature map $\mathbf{h}(\cdot)$, the low-dimensional response surface in feature space $f_{Z}$, and the reconstruction mapping $\mathbf{g}(\cdot)$.

\subsection{Manifold Gaussian processes for response surface learning in feature space}
\label{sec:ManifoldGP}

We expect the response surface to predict the value of the black-box objective function $f_X$ with calibrated uncertainty associated with each prediction. 
GPs are probabilistic models that allow for an analytic computation of posterior predictive function values within a Bayesian framework, and they are the standard model in BO for modeling the response surface. 

A GP is a distribution over functions $f_{Z}\sim\mathcal{GP}(m(\cdot), k(\cdot, \cdot))$ and is fully specified by a \emph{mean function} $m\colon\mathcal{Z}\rightarrow\mathbb{R}$, and a \emph{covariance function\slash kernel} $k\colon\mathcal{Z}\times\mathcal{Z}\rightarrow\mathbb{R}$. The kernel computes the covariance between pairs of function values as a function of the corresponding inputs, i.e. $\text{Cov}(f_{Z}(\mathbf{z}), f_{Z}(\mathbf{z}'))~=~k(\mathbf{z}, \mathbf{z}')$, and thereby encodes regularity assumptions about $f_{Z}$, such as smoothness or periodicity. Common kernel choices in the BO literature include the \emph{squared exponential} and \emph{Mat\'ern}  kernels \cite{Frazier2018}.

In our feature space optimization, we phrase lines \ref{h_map}--\ref{f_Zmap} of Algorithm~\ref{FeatureBO} as a single learning problem. Therefore, we need a GP that learns useful representations $\vec z$ of inputs $\vec x$ for the regression task together with $f_{Z}$. A manifold GP (MGP) \cite{ManifoldGP2016,DeepK2016} addresses this issue by composing two mappings: The deterministic feature map $\mathbf{h}$ with parameters $\boldsymbol{\theta}_{h}$ and a GP $f_Z\sim\mathcal{GP}(m, k)$ with kernel hyper-parameters $\boldsymbol{\theta}_{k}$. The GP  models the relationship between features $\vec z$ and function values $y\in\mathbb{R}$ in observation space. The resulting composite model $f_X := f_Z\circ \mathbf{h}$ is a GP so that $f_{X}\sim\mathcal{GP}(m_{M}, k_{M})$ with mean function given by $m_{M}(\mathbf{x}) = m(\mathbf{h}(\mathbf{x}))\label{m_m}$ and the covariance function given by $k_{M}(\mathbf{x}, \mathbf{x}')  = k(\mathbf{h}(\mathbf{x}), \mathbf{h}(\mathbf{x}'))$ respectively. Given high-dimensional training inputs $\mathbf{X}$ and corresponding observations $\vec y$ of the objective function, we find model parameters $\{\boldsymbol{\theta}_{h},\boldsymbol{\theta}_{k}\}$ that maximize the marginal likelihood (evidence) $\{\boldsymbol{\theta}_{{h}}^{*},\,\boldsymbol{\theta}_{k}^{*}\} \in \underset{\boldsymbol{\theta}_{{h}},\boldsymbol{\theta}_{k}}{\arg\max}\, p(\mathbf{y}|\mathbf{X}, \boldsymbol{\theta}_{{h}},\boldsymbol{\theta}_{k})$.
This objective allows us to learn a low-dimensional embedding as a by-product of the supervised GP regression. 

Unsupervised dimensionality reduction usually solves an orthogonal task to that of learning a response surface. Algorithms, such as PCA or variational auto-encoders \cite{JimenezRezende2014a}, achieve compact data representations by optimizing objectives that are not necessarily useful in a supervised setting \cite{Wahlstrom2015}.
The MGP, instead, leads to low-dimensional representations that are optimal (locally) for the regression task at hand.

We use a multi-layer feed-forward neural network with sigmoid activation functions as a feature map (encoder) $\vec h$, resulting in a feature space $\mathcal{Z}=[0,1]^{d}$. Neural networks as an explicit feature map within an MGP have already been applied successfully for modeling non-smooth responses in robot locomotion \cite{ManifoldGP2016,Cully2015}. Deep networks have also proven useful for learning the orientation of images from high-dimensional images \cite{DeepK2016}. With a Gaussian likelihood, the MGP posterior predictive distribution at a test point $\mathbf{x}_{\star}\in\mathcal X$ is Gaussian distributed with mean and variance given by
\begin{align}
\begin{split}
    \mathbb{E}[f_X(\mathbf{x}_{\star})] &= m_{M}(\mathbf{x}_{\star}) + k_{M}(\mathbf{x}_{\star}, \mathbf{X})\mathbf{K}_{My}^{-1}(\mathbf{y} - m_{M}(\mathbf{X})) \\
    &= m(\mathbf{z}_{\star}) + k(\mathbf{z}_{\star}, \mathbf{Z})\mathbf{K}_{y}^{-1}(\mathbf{y} - m(\mathbf{Z})) \label{post_meanGP}
    \end{split}
    \\
    \begin{split}
    \mathbb{V}[f_X(\vec x_{\star})]&= k_{M}(\mathbf{x}_{\star}, \mathbf{x}_{\star}) - k_{M}(\mathbf{x}_{\star},\mathbf{X})\mathbf{K}_{My}^{-1}k_{M}(\mathbf{X}, \mathbf{x}_{\star}) \\
    &= k(\mathbf{z}_{\star}, \mathbf{z}_{\star}) - k(\mathbf{z}_{\star},\mathbf{Z})\mathbf{K}_{y}^{-1}k(\mathbf{Z}, \mathbf{z}_{\star}),
    \end{split}\label{post_covGP}
\end{align}
respectively, with  $\mathbf{z}_{\star}:=\mathbf{h}(\mathbf{x}_{\star})$ and $\mathbf{Z}:=\mathbf{h}(\mathbf{X})$. Moreover, $k_{M}(\mathbf{x}_{\star}, \mathbf{X}) = k(\mathbf{z}_{\star}, \mathbf{Z})=[k(\mathbf{z}_{\star}, \mathbf{z}_{i})]_{i=1}^{N}$, where $N$ is the size of the training dataset, $\mathbf{K}_{My}:=k_{M}(\mathbf{X}, \mathbf{X}) + \sigma_{n}^{2}\mathbf{I}$, $\mathbf{K}_{y}:=k(\mathbf{Z}, \mathbf{Z}) + \sigma_{n}^{2}\mathbf{I}$, $k_{M}(\mathbf{X}, \mathbf{X}):=k(\mathbf{Z}, \mathbf{Z})$, and $ m_{M}(\mathbf{X}):=m(\mathbf{Z})=[m(\mathbf{z}_{i})]_{i=1}^{N}$ computes the prior mean function evaluated at the embedded training inputs $\vec Z$. 
Posterior predictions can be computed using both the feature and data space. Equations (\ref{post_meanGP})--(\ref{post_covGP}) appear in the definition of the acquisition functions in  (\ref{equationPI})--(\ref{equationUCB}) as mean $\mu(\vec x):=\mathbb{E}[f_X(\mathbf{x})]$ and standard deviation $\sigma(\vec x):=\sqrt{\mathbb{V}[f_X(\vec x)]}$ of the posterior predictions of the surrogate model.


The MGP defines a GP on $\mathcal{X}$, but allows us to learn a response surface in the lower-dimensional feature space $\mathcal Z$. This is key for optimizing the acquisition function in a low-dimensional space $\mathcal Z$ instead of the original data\slash parameter space $\mathcal X$. Thus far, we have detailed the feature-space BO procedure up to line \ref{acq_max} in Algorithm~\ref{FeatureBO}. Once we found an optimizer $\vec z_*$ of the acquisition function, we need to project it back into the original data space $\mathcal X$ in order to evaluate the true objective $f_{X}$, whose domain is $\mathcal X$. This can be done by means of a reconstruction mapping (decoder), which we detail in the following.




\subsection{Input reconstruction with manifold multi-output Gaussian processes}
\label{sec:mMOGP}
\label{sec:MOGP}
In the following, we present the reconstruction part (decoder) of our feature space optimization model described in Figure~\ref{fig:GP-AE}.
We are interested in modeling the functional relationship between the feature space $\mathcal{Z}$ and the data space ${\mathcal{X}}$ for step \ref{reconstruct} in Algorithm~\ref{FeatureBO}, which requires us to evaluate $f_X$. We therefore consider a vector-valued function $\mathbf{g}=\{g_{i}\}_{i=1}^{D}$, where each component $g_{i}\colon\mathcal{Z}\rightarrow{\mathcal{X}}_{i}$ maps vectors in feature space to the $i$-th coordinate of high-dimensional data, i.e. $g_i(\mathbf{z}) = \tilde{{x}}^{(i)}\in\mathcal X_{i}$. Multi-output GPs (MOGPs) \cite{AlvarezLawrence,MOGP_0,MOGP_1,RegressionNetworks2011,alvarez2009sparse,osborne2008towards,seeger2005semiparametric,boyle2005dependent} define a prior over vector-valued functions and explicitly allow for output correlations. An MOGP $\mathcal{GP}(\mathbf{m}, \mathbf{K})$ is fully specified by a mean vector function $\mathbf{m}\colon\mathcal{Z}\rightarrow\mathbb{R}^{D}$ and a positive, semi-definite matrix-valued covariance function $\mathbf{K}\colon\mathcal{Z}\rightarrow\mathbb{R}^{D\times D}$, which computes the correlation between observations in the same output coordinate and cross-correlations between the $D$ different outputs.

Here we consider the \emph{intrinsic coregionalization model} (ICM) \cite{ICM1997,ICM2013}, which structures the covariance matrix as a Kronecker product.
This model is particularly suitable for trading off number of model parameters and expressiveness of the vector valued function. In particular, the ICM facilitates information sharing across different tasks by adopting the same covariance function. It has been successfully adopted in robotics for learning inverse dynamics \cite{Williams2009}. Hence, this model requires fewer parameters than the linear model of coregionalization \cite{AlvarezLawrence} and allows for exploiting properties of the Kronecker product for efficient training and prediction. 

In our reconstruction model, we  need to ensure that the output space is exactly $\mathcal{X}$. If we start from a data space $\mathcal X = [0,1]^{D}$ the reconstructions need to belong to this hypercube. This property is not guaranteed by the MOGP. In order to satisfy this constraint, we consider a strictly monotonic output squashing function $\Psi$, as introduced in the context of warped GPs~\cite{snelson2004warped}. This allows us to define a corresponding inverse transformation $\Psi^{-1}$ that is applied to the data in input to the model. The resulting output of the MOGP at test time is then squashed through the transformation $\Psi$. Since the reconstruction of the MOGP is a distribution $p(\tilde{\vec {x}}_{\star}|\vec X,\tilde{\vec X},\vec z_{\star})$, we evaluate the expectation with respect to this distribution of the transformed outputs, i.e. $\vec x_{t+1}=\mathbb{E}_{p}[\Psi(\tilde{\vec x}_{\star})|\vec X,\tilde{\vec X},\vec z_{\star}]$. In this paper, we choose the Gaussian cumulative density function as a monotonic squashing function $\Psi :=\Phi$ for warping the outputs of our reconstruction model~\cite{snelson2004warped}. The motivation for this choice is twofold: the inverse mapping $\Psi^{-1}$ is defined as the Probit function, which is a well known function, and the expectation $\mathbb{E}_{p}[\Psi(\tilde{\vec x}_{\star})|\vec X,\tilde{\vec X},\vec z_{\star}]$
 with respect to the distribution $p(\tilde{\vec {x}}_{\star}|\vec X,\tilde{\vec X},\vec z_{\star})$ at test reconstructions can be derived analytically~\cite{Rasmussen2006}.

\textbf{Intrinsic coregionalization model.} The ICM \cite{ICM1997,ICM2013} applies a linear mapping to a set of latent functions. In particular, we consider a set of $P$ latent functions $u_{i}\colon\mathcal{Z}\rightarrow\mathbb{R}$, that are assumed to be \emph{sample paths}, i.e. sample functions independently drawn from the same GP prior $\mathcal{GP}(m_{c}, k_{c})$. The ICM model expresses the vector-valued function as a linear combination of these sample functions
$\mathbf{g}(\mathbf{z})=\mathbf{A}\mathbf{u}(\mathbf{z})$, where $\mathbf{u}(\mathbf{z})\in\mathbb{R}^{P}$ is the collection of the $P$ sample paths' evaluations at  $\mathbf{z}$, and $\mathbf{A}\in\mathbb{R}^{D\times P}$ is the linear mapping that couples the independent vector and parameterizes the ICM model. As a result, $\mathbf{g}$ is an MOGP $\mathcal{GP}(\mathbf{m}, \mathbf{K})$ with mean function $\mathbf{m}=\mathbf{A}\mathbf{m}_{c}$, where $\mathbf{m}_{c}=[m_{c}]_{i=1}^{P}$ is obtained by repeating the single-valued mean function $m_{c}$ in a $P$-vector. The covariance function is  $\mathbf{K}(\mathbf{z}, \mathbf{z}') = \mathbf{A}\mathbf{A}^{T}\otimes k_{c}(\mathbf{z}, \mathbf{z}')$, 
where $k_{c}$ is the covariance function for the GP prior, $\otimes$ is the Kronecker product and the matrix $\mathbf{A}\mathbf{A}^{T}$ is denoted as the coregionalization matrix. Note that $k_c$ may differ from the covariance function $k$ used for the response surface $f_Z$.

\textbf{Reconstruction Model.} For the reconstruction task in line \ref{reconstruct} of Algorithm~\ref{FeatureBO}, we introduce the manifold MOGP with intrinsic coregionalization model (mMOGP), which shares the feature map $\mathbf{h}$ with the MGP used for learning the response surface; see Section \ref{sec:ManifoldGP}. Without loss of generality, we assume a prior zero-mean vector function for the mMOGP $\mathcal{GP}(\mathbf{0}, \mathbf{B}\otimes k_{MO})$, where $k_{MO}(\mathbf{x}, \mathbf{x}') = k_{c}(\mathbf{h}(\mathbf{x}), \mathbf{h}(\mathbf{x}'))$ and the matrix $\mathbf{B}=\mathbf{AA}^{T}$. 
We can interpret this model as an auto-encoder, where the MGP $\mathbf{g}\circ \mathbf{h}:\mathcal X \to\mathcal Z$ plays the role of the encoder, and the MOGP the role of the decoder, mapping low-dimensional features back into data space.



\subsection{Joint training}
\label{sec:joint_training}
The joint training of the MGP, which models the response surface, and the mMOGP, which is used for the reconstruction (see also  Figure \ref{fig:GP-AE}), is performed by maximizing a rescaled version of the log-marginal likelihood 
\begin{align}
    \mathcal{L} \propto &-\mathbf{y}^T\mathbf{K}_{y}^{-1}\mathbf{y}-\log |\mathbf{K}_{y}|
    -\dfrac{1}{D}\left(\mathbf{x}_{V}^T\mathbf{K}_{V}^{-1}\mathbf{x}_{V} + \log |\mathbf{K}_{V}|\right) + \text{const}.
    \label{eq:marginal likelihood}
\end{align}
Here, $\mathcal{L}$ comprises terms from both the MGP and mMOGP models,
where $\mathbf{K}_{y}$ is defined in  (\ref{post_meanGP}), and the covariance matrix of the mMOGP $\mathbf{K}_{V}=\bar{\mathbf{K}} + \sigma_{n}^{2}\mathbf{I}$ is obtained by evaluating the Kronecker product $\bar{\mathbf{K}}=\mathbf{B}\otimes k_{c}(\mathbf{Z}, \mathbf{Z})$ with the mMOGP kernel $k_{c}$. The vector $\mathbf{x}_{V}$ is a concatenation of the columns of the data $\mathbf{X}$. 
The maximizers $[\boldsymbol{\theta}_{h}^{*},\boldsymbol{\theta}_{k}^{*},\boldsymbol{\theta}_{c}^{*}]$ of the log-marginal likelihood are the parameters $\boldsymbol{\theta}_{h}^{*}$ of the feature map $\mathbf{h}$ (which is shared between the MGP and the mMOGP), the hyper-parameters $\boldsymbol{\theta}_{k}^{*}$ of the kernel $k$ and the hyper-parameters $\boldsymbol{\theta}_{c}^{*}$ of $k_{c}$ including the coregionalization matrix $\vec B$ for the mMOGP, respectively. The rescaling factor $1/D$ balances the contributions of the two log-marginal likelihood terms involved in  training. The dimensions of the matrix $\mathbf{K}_{V}$ are $ND\times ND$ which correspond to repeating the $\vec K_{y}$ matrix $D$ times in a block-diagonal fashion. This block diagonal would then have quadratic form equal to $D\vec y^{T}\vec K_{y}^{-1}\vec y$ and log determinant equal to $D\log |\mathbf{K}_{y}|$. Thus, an equivalent rescaling is to divide the reconstruction terms $\mathbf{x}_{V}^T\mathbf{K}_{V}^{-1}\mathbf{x}_{V}$ and $\log |\mathbf{K}_{V}|$ by $D$. Optimization of~\eqref{eq:marginal likelihood} is performed via gradient-based methods \cite{Lu1994,Zhu1997AlgorithmOptimization}.

Modeling the black-box objective function $f_X$ is orthogonal to the reconstruction problem. However, when training these tasks jointly, they have a regularization effect on the optimization of the parameters $\boldsymbol{\theta}_{h}$ of the feature embedding in the sense that the mapping $\mathbf{h}$ will not overfit to a single regression task: the parameters $\boldsymbol{\theta}_{{h}}$ will give rise to a feature space embedding that is useful for both the modeling of the objective and the reconstruction of the original inputs.

The major computational bottleneck for evaluating the marginal likelihood comes from the term $\mathbf{x}_{V}^T\mathbf{K}_{V}^{-1}\mathbf{x}_{V}$, which requires inverting an $ND\times ND$ covariance. We reduce the computational complexity of this operation to $\mathcal{O}(N^{3}) + \mathcal{O}(D^{3})$ by exploiting the properties of the Kronecker product, tensor algebra \cite{TensorAL1999} and structured GPs \cite{Gilboa2015,Saatci2012} as shown in the following section. 

\subsection{Computationally efficient mMOGP}
\label{sec:comp_efficient_model}

For the reconstruction mapping $\mathbf{g}$, we use the posterior mean of the mMOGP with intrinsic coregionalization model and apply exact inference and training via rescaled marginal likelihood maximization. While the ICM enables modeling correlation between arbitrary pairs of dimensions, it also requires computing  a Kronecker product to evaluate the full covariance matrix of all outputs
\begin{equation}
\bar{\mathbf{K}} = \mathbf{B}\otimes k_{c}(\mathbf{Z}, \mathbf{Z}),\label{Kronecker_product}
\end{equation}
where $k_{c}(\mathbf{Z}, \mathbf{Z})$ is the covariance matrix obtained from the training inputs $\mathbf{Z}$ in feature space and the $\mathbf{B}$ matrix is the coregionalization matrix of the ICM. Inverting the full covariance matrix $\bar{\mathbf{K}}$ requires $\mathcal{O}(N^{3}D^{3})$ and easily becomes intractable in high-dimensional spaces even for small $N$. Storing this full covariance matrix  required $\mathcal{O}(N^{2}D^{2})$ space and also becomes challenging in high dimensions. For an efficient implementation of the ICM, we exploit properties of the Kronecker product and apply results from structured GPs \cite{Gilboa2015,Saatci2012} that allow for efficient training and predictions in $\mathcal{O}(N^{3}) + \mathcal{O}(D^{3})$ time and $\mathcal{O}(ND)$ space. In particular, we are interested in the full covariance matrix under the assumption of a Gaussian likelihood for the multi-output observations, i.e. $\bar{\mathbf{K}} + \sigma_{n}^{2}\mathbf{I}$. We first express the full covariance matrix in terms of its eigendecomposition, i.e. $\bar{\mathbf{K}} = \mathbf{Q}\boldsymbol{\Lambda}\mathbf{Q}^{T}$. This allows expressing the inverse of the covariance from noisy targets as
\begin{equation}
\left(\bar{\mathbf{K}} + \sigma_{n}^{2}\mathbf{I}\right)^{-1} = \mathbf{Q}\left(\boldsymbol{\Lambda} + \sigma_{n}^{2}\mathbf{I}\right)^{-1}\mathbf{Q}^{T},\label{InvQLQ}
\end{equation}
where both $\boldsymbol{\Lambda}$ and $\sigma_{n}^{2}\mathbf{I}$ are diagonal and can be trivially inverted. However, the eigendecomposition of an $ND\times ND$ matrix would still be cubic in the product between the number of dimensions and the number of data. By the properties of the Kronecker product, we can express  the eigendecomposition itself with a Kronecker structure, i.e.
\begin{equation}
    \bigotimes\nolimits_{l=1}^{W}\,\mathbf{K}_{l} = \bigotimes\nolimits_{l=1}^{W}\,\mathbf{Q}_{l}\bigotimes\nolimits_{l=1}^{W}\,\boldsymbol{\Lambda}_{l}\Big(\bigotimes\nolimits_{l=1}^{W}\,\mathbf{Q}_{l}\Big)^{T},\label{KronQLQT}
\end{equation}
where each term of the Kronecker product on the left-hand side $\mathbf{K}_{l}\in\mathbb{R}^{G_{l}\times G_{l}}$ has eigendecomposition $\mathbf{K}_{l}=\mathbf{Q}_{l}\boldsymbol{\Lambda}_{l}\mathbf{Q}_{l}^{T}$ for $l=1,...,W$, where $W$ is number of factors in the Kronecker product. In our ICM model $W=2$, because  the coregionalization matrix $\mathbf{B}$ Kronecker multiplies $k_{c}(\vec Z, \vec Z)$, the covariance matrix of the observations; see (\ref{Kronecker_product}). Thus, from  (\ref{InvQLQ})--(\ref{KronQLQT}), we are allowed to invert the covariance from noisy targets by separately decomposing the covariance matrix $k_{c}(\mathbf{Z},\mathbf{Z})=\mathbf{Q}_{k}\boldsymbol{\Lambda}_{k}\mathbf{Q}_{k}^{T}$ and the coregionalization matrix $\mathbf{B}=\mathbf{Q}_{b}\boldsymbol{\Lambda}_{b}\mathbf{Q}_{b}^{T}$, which require $\mathcal{O}(N^{3})$ and $\mathcal{O}(D^{3})$ time, respectively; see line~\ref{eigh} of Algorithm~\ref{mv_fast}.

\begin{algorithm}
\caption{Efficient computation of the inverse for matrices that have a Kronecker structure and spherical additive noise. Subroutine \texttt{matvecmul}: fast matrix-vector multiplication for matrices that can be expressed as a Kronecker product. Here the function \texttt{eigh} returns the eigen-decomposition of a matrix.}\label{mv_fast}
  \begin{algorithmic}[1]
   	\State {\bfseries Input matrices:} $\{\mathbf{K}_{l}\in\mathbb{R}^{G_{l}\times G_{l}}\}_{l=1}^{W}$
   	\State {\bfseries Input vector:} $\mathbf{x}_{V}\in\mathbb{R}^{N_{V}}, \ \ N_{V} = \prod_{l=1}^{W}\,G_{l}$
   	\State {\bfseries Input variable:} $\sigma_{n}^{2}$
      \For{$l = 1,2,...,W$}
        \State $ \boldsymbol{\Lambda}_{l}, \mathbf{Q}_{l}, = \texttt{eigh}(\vec K_{l})$\label{eigh}
        \Comment{Eigen-decomposition of each input matrix}
        \EndFor
        \State  $\mathbf{s} = \texttt{matvecmul}(\bigotimes_{l=1}^{W}\,\mathbf{Q}_{l}^{T}, \mathbf{x}_{V})$ 
        \Comment{Fast matrix-vector multiplication}
        \State  $\mathbf{D} = \bigotimes_{l=1}^{W}\,\boldsymbol{\Lambda}_{l} + \sigma_{n}^{2}\vec I$ 
        \Comment{Diagonal term with eigenvalues and noise}
        \State $\mathbf{w} = \vec D^{-1} \vec s = [s_{i}/D_{i, i}]_{i=1}^{N_{V}}$ \Comment{Standard matrix-vector multiplication}
      \State \textbf{Return} $\vec r = \texttt{matvecmul}\left(\bigotimes_{l=1}^{W}\,\mathbf{Q}_{l}, \mathbf{w}\right) = \left(\bigotimes_{l=1}^{W}\,\mathbf{K}_{l} + \sigma_{n}^{2}\vec I \right)^{-1}\vec x_{V}$
      \Comment{Fast matrix vector multiplication of an inverse with Kronecker structure and a vector}
  \end{algorithmic}
  
    \begin{algorithmic}[1]
   	\State {\textbf{Procedure} \texttt{matvecmul}($ \bigotimes_{l=1}^{W}\,\mathbf{K}_{l}, \mathbf{x}$) }
   	\State {\bfseries Input matrices:} $\{\mathbf{K}_{l}\in\mathbb{R}^{G_{l}\times G_{l}}\}_{l=1}^{W}$
   	\State {\bfseries Input vector:} $\mathbf{x}\in\mathbb{R}^{N_{V}}, \ \ N_{V} = \prod_{l=1}^{W}\,G_{l}$
   	\State  $\mathbf{r}=\mathbf{x}$ \Comment{Initialize result}
      \For{$l = W,W-1,...,1$}
        \State  $\mathbf{R} = \texttt{reshape}(\mathbf{r},\,[G_{l}, N_{V}/G_{l}])$ \Comment{Reshape results}
        \State $\mathbf{Z} = \mathbf{K}_{l}\mathbf{R}$
        \Comment{Matrix-tensor product }
        \State  $\mathbf{r} = \texttt{vec}(\mathbf{Z}^\top)$ \Comment{Reshape results}
      \EndFor
      \State \textbf{Return} $\mathbf{r} = \left(\bigotimes_{l=1}^{W}\,\mathbf{K}_{l}\right)\mathbf{x}$
      \Comment{Fast matrix vector multiplication for matrix with Kronecker structure}
  \end{algorithmic}
\end{algorithm}

Storing this inverse matrix and multiplying it by a vector still requires $\mathcal{O}(N^{2}D^{2})$ space and run time, respectively, so that this step becomes the main bottleneck for efficient mMOGP training and predictions computation. To address this issue we represent the expensive matrix-vector multiplication as a sequence of small matrix-tensor multiplications without computing the full Kronecker product  \cite{TensorAL1999}. In particular, we are interested in efficiently evaluating
\begin{equation}
    \mathbf{r} = \left(\bigotimes\nolimits_{i=1}^{W}\,\mathbf{K}_{i}\right)\mathbf{x}.
\end{equation}
We first represent the multiplication of a matrix with Kronecker structure by a vector as a {tensor product}. A \emph{tensor} $\mathbf{T}_{i_{1},...,i_{V}}$ can be interpreted as an extension of matrices to objects where elements are indexed using a set of $V$ indices: $i_{1},...,i_{V}$, where the number $V$ is referred to as the \emph{order} of the tensor.
With the definition of the Kronecker product we express the left-hand side of~(\ref{KronQLQT}) as a tensor
\begin{align}
    \Big[\bigotimes\nolimits_{l=1}^{W}\,\mathbf{K}_{l}\Big]_{i, j} &= \left[\mathbf{K}_{1}\right]_{i_{1}, j_{1}}\cdot\,...\,\cdot\left[\mathbf{K}_{W}\right]_{i_{W}, j_{W}}\label{tensorKron},\\
    1\leq i_{l}&,j_{l}\leq G_{l}\,,\quad
    1\leq i,j\leq \prod\nolimits_{l=1}^{W}G_{l}.\nonumber
\end{align}
The right-hand side of (\ref{tensorKron}) coincides with a tensor $\mathbf{T}^{K}_{i_{1},j_{1},...,i_{W},j_{W}}$, and a similar tensor-representation can be obtained for the $\prod_{l=1}^{W}G_{l}$-long vector $\mathbf{x}$, i.e. $\mathbf{T}^{X}_{j_{W},...,j_{1}}$.
A tensor product between the tensors $\mathbf{T}^{K}_{i_{1},j_{1},...,i_{W},j_{W}}$ and $\mathbf{T}^{X}_{j_{W},...,j_{1}}$ applies a contraction along the indices of the second tensor, i.e.
\begin{equation}
    \sum\nolimits_{j_{1}}...\sum\nolimits_{j_{W}}\,\mathbf{T}^{K}_{i_{1},j_{1},...,i_{W},j_{W}}\mathbf{T}^{X}_{j_{W},...,j_{1}}.
\end{equation}
This tensor contraction can be expressed in terms of a sequence of {tensor-transposed} matrix-tensor products
\begin{align}
    \Big(\bigotimes\nolimits_{l=1}^{W}\,\mathbf{K}_{l}\Big)\mathbf{x} &= \text{vec}\Big(\left(\mathbf{K}_{1}\cdots\left(\mathbf{K}_{W}\mathbf{T}^{X}\right)^{\top}\right)^{\top}\Big)\label{seqMatTens}
\end{align}
with $\mathbf{K}_{l}\mathbf{T}^{X} = \sum_{k=1}^{G_{l}}\,\left[\mathbf{K}_{l}\right]_{i_{1}, k}\mathbf{T}^{X}_{k, j_{2},...,j_{W}}$. The function $\text{vec}(\cdot)$ returns the vectorized form of a matrix by stacking its columns vertically. The tensor transposition $\top$ applies a cyclic permutation to the order of the indices in a tensor. As a result, the right-hand side in (\ref{seqMatTens}) allows us to evaluate the expensive matrix-vector product without computing and storing the Kronecker product. Algorithm~\ref{mv_fast} shows the main steps of the efficient matrix inversion and matrix-vector multiplication for matrices that feature a Kronecker structure. The matrix vector multiplication subroutine is expressed as a sequence of tensor-transpose matrix-tensor products.

\section{Constrained acquisition}

We defined a joint probabilistic model for the response surface learning and the input reconstruction tasks; see lines \ref{macro_i}--\ref{f_Zmap} and \ref{reconstruct} of Algorithm~\ref{FeatureBO}, respectively. We are now concerned with the maximization of the acquisition function in feature space; see line \ref{acq_max} of Algorithm~\ref{FeatureBO}. We aim at maximizing the acquisition function in a low-dimensional feature space of the original data\slash parameter space $\mathcal X$. However, one problem that arises with the mMOGP decoding is that locations in feature space, which are too far away from data, will be mapped back to the mMOGP prior. 
Since the acquisition function is a key driver of exploration in BO, this is a problem. We address this limitation by introducing a constraint based on the Lipschitz continuity of the mMOGP posterior. This will ensure that candidates $\vec z_{*}\in\mathcal Z$ selected in feature space will not collapse to the origin $\mathbf{0}\in\mathbb{R}^{D}$ if the reconstruction is defined as $\tilde{\vec x}_{*}=\boldsymbol{\mu}(\vec z_{*})$, where $\boldsymbol{\mu}$ is the posterior mean of the mMOGP.
\begin{figure*}
    \centering
    \subfigure{\includegraphics[width=0.9\linewidth]{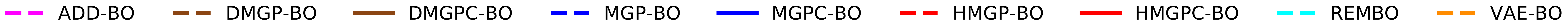}}
    \setcounter{subfigure}{0}
    \subfigure[EI]{\includegraphics[width=.3\linewidth]{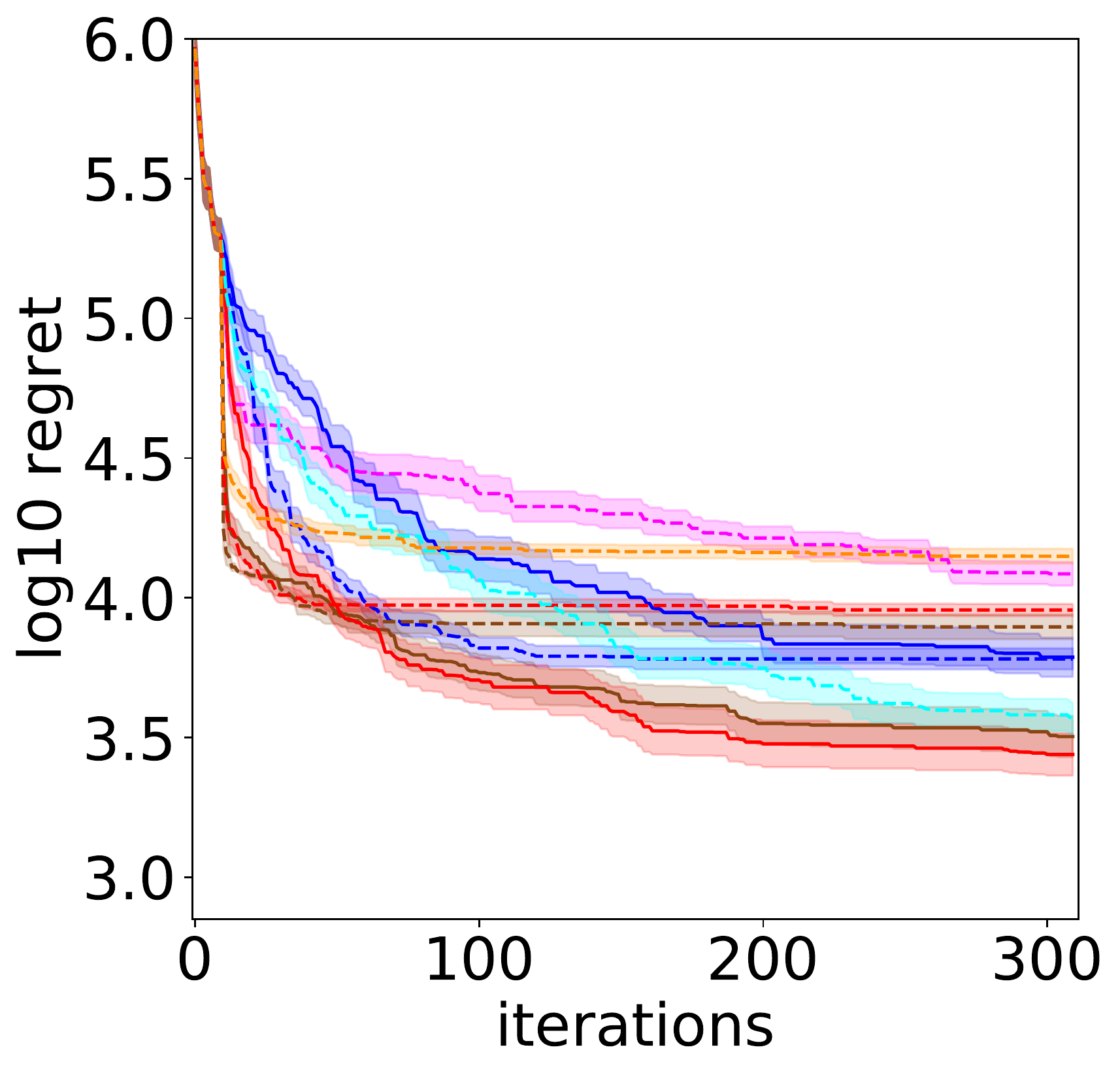}\label{fig:RosLinearEI}}%
    \hfill
    \subfigure[UCB]{\includegraphics[width=.3\linewidth]{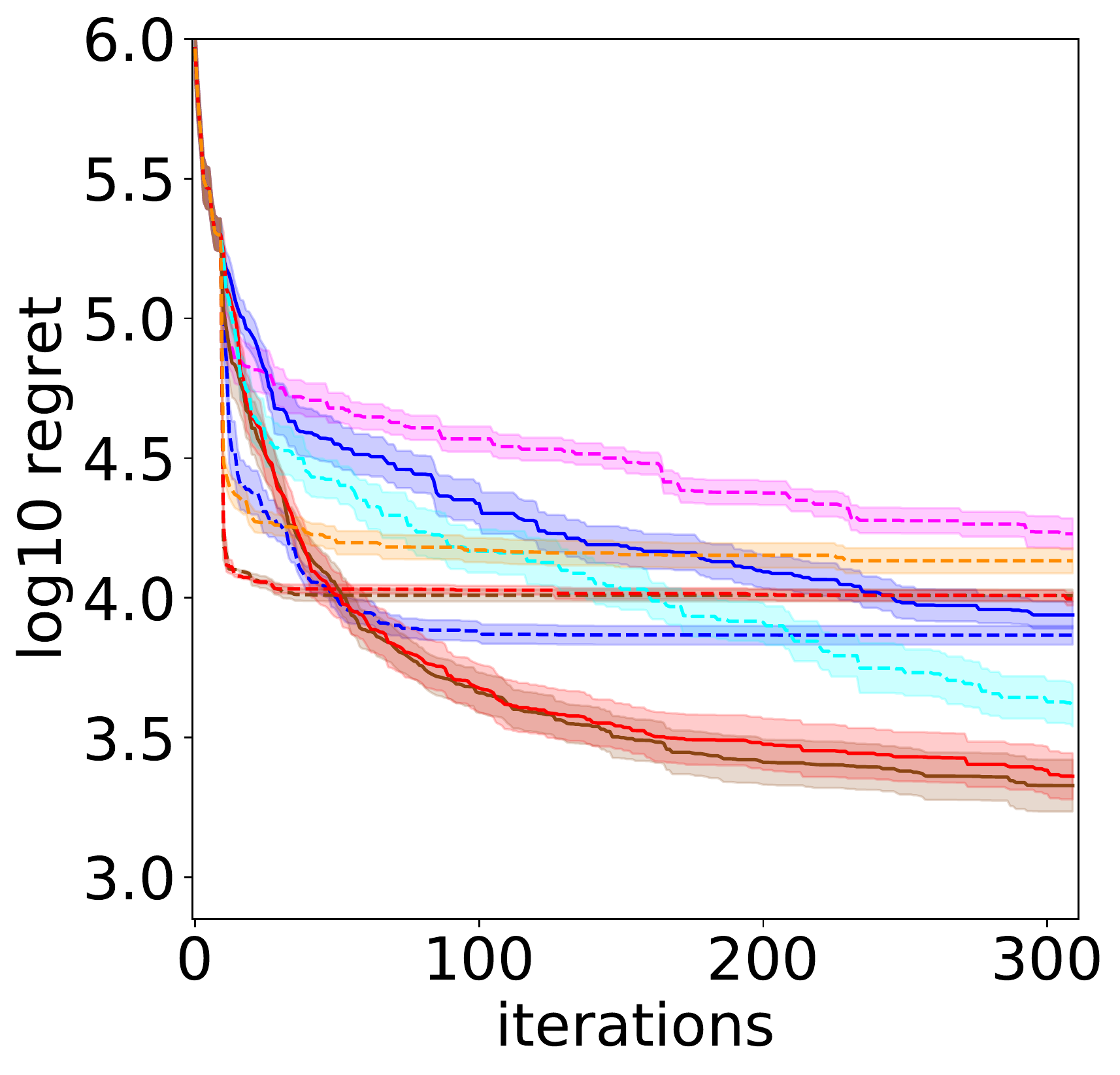}\label{fig:RosLinearUCB}}%
    \hfill
    \subfigure[PI]{\includegraphics[width=.3\linewidth]{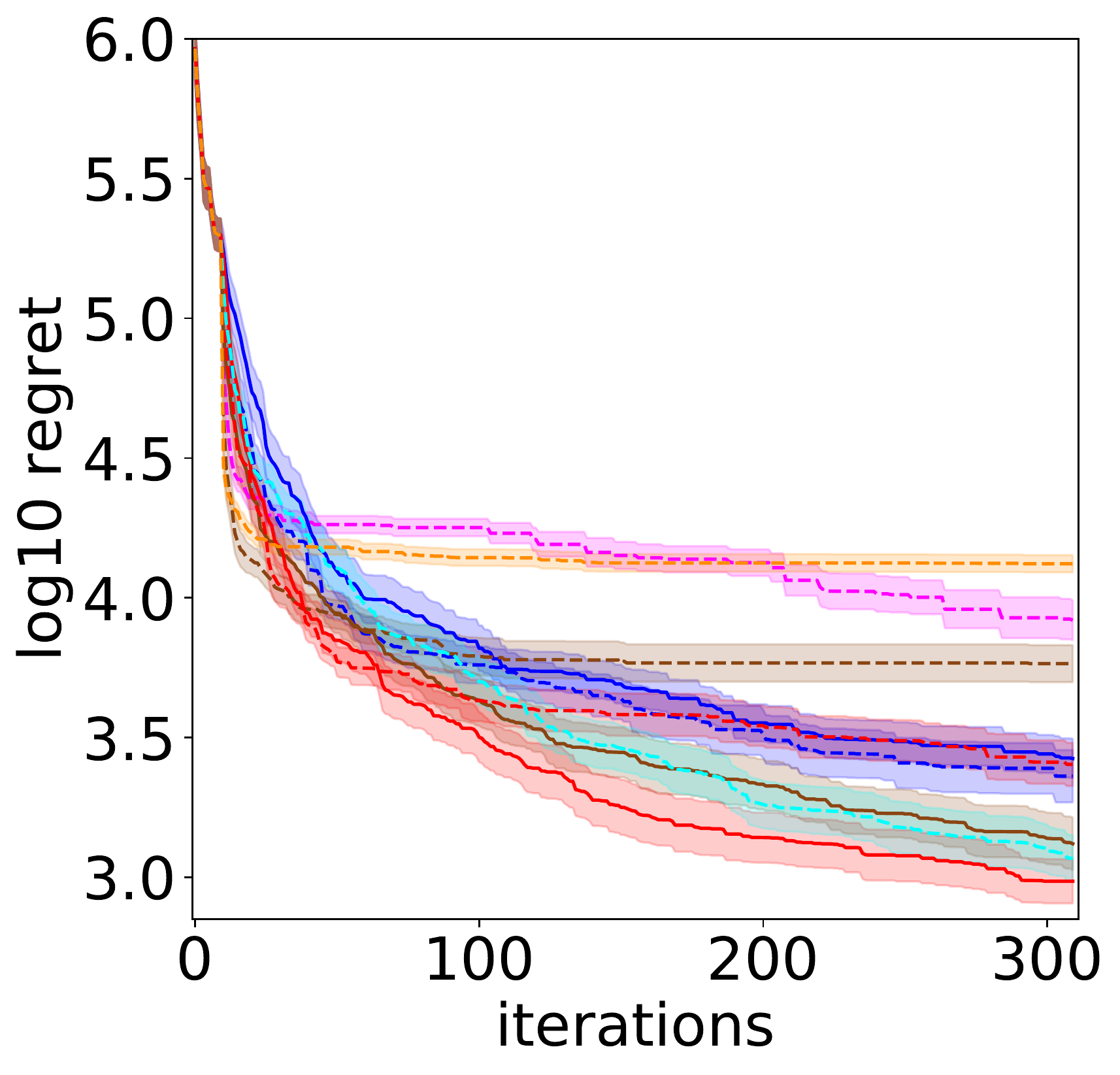}\label{fig:RosLinearPI}}
    
    \caption{Results with Rosenbrock objective function of BO in feature space. The objective function is characterized by a linear embedding to reach $D=60$ dimensions. Baselines MGPC-BO, HMGPC-BO and DMGPC-BO (solid) apply nonlinearly constrained acquisition maximization and recover no worse regret at termination than the unconstrained versions MGP-BO, HMGP-BO and DMGP-BO.}\label{fig:RosLinear}
    
\end{figure*}

We want to leverage information from observed data for the multi-output mapping and exploit it when optimizing the acquisition function in feature space. This can be achieved by upper-bounding the Euclidean distance

\begin{equation}
\text{dist}(\mathbf{z}, \mathbf{Z}_{t}) = \underset{1\leq i\leq N_{t}}{\min}\,\|\mathbf{z}_{i}-\mathbf{z}\|_{2}\label{distMin}
\end{equation}
in feature space
 between the optimization variable $\mathbf{z}$ and the embedded training data $\mathbf{Z}_{t}=[\mathbf{z}_{1},...,\mathbf{z}_{N_{t}}]$. Here, $N_{t}$ is the number of data points available at BO iteration $t$. The desired upper bound is obtained by exploiting the Lipschitz continuity property of the multi-output posterior mean for which 
\begin{equation}
    |[\boldsymbol{\mu}(\mathbf{z})]_{i} - [\boldsymbol{\mu}(\mathbf{z}')]_{i}|\leq L \|\mathbf{z}-\mathbf{z}'\|.
\end{equation}
Here, $L$ denotes the Lipschitz constant of the posterior mean $\boldsymbol{\mu}$ of the mMOGP. For common kernels, such as Mat\'ern$_{52}$ and squared exponential, the posterior mean is Lipschitz continuous. The upper bound
\begin{equation}
    \text{dist}(\mathbf{z}, \mathbf{Z}_{t})\leq {\mu_{\max}(\mathbf{z}^{*})}/L\label{Lconstraint}
\end{equation}
allows us to specify how far from the data we can move in feature space without falling back to the prior on all coordinates of the reconstruction. Here $\mathbf{z}^{*}$ minimizes the distance in~(\ref{distMin}), while the numerator on the right-hand side is the component-wise maximum of $\boldsymbol{\mu}(\mathbf{z}^{*})$.
%
%
We estimate the Lipschitz constant as the maximum norm of the Jacobian of the posterior mean of the mMOGP \cite{GonzalezBatchBO2016}
\begin{equation}
    L = \underset{\mathbf{z}\in\mathcal{Z}}{\max}\,\|\nabla_{\vec z} \boldsymbol{\mu}(\mathbf{z})\|\label{GPL-CA}.
\end{equation}
This maximization returns a valid Lipschitz constant \cite{GonzalezBatchBO2016} for the multi-output mapping for any choice of norm in  (\ref{GPL-CA}). The Jacobian of the posterior mean is represented by a $D\times d$ matrix and we adopt the max norm $\|\nabla_{\vec z}\boldsymbol{\mu}(\mathbf{z})\|_{\infty}=\max |\mu_{i,j}^{'}|$ for $i=1,...,D$ and $j=1,...,d$. 
Lower values of valid Lipschitz constants $L$ allow for exploration in larger regions of the feature space that still satisfy the nonlinear constraint in  (\ref{Lconstraint}).

\section{Experiments}\label{results}

We report results on a set of high-dimensional benchmark functions that possess an intrinsic low dimensionality. In particular, we (i) assess the benefits of adopting a model structure as presented in Figure \ref{fig:GP-AE}; (ii) analyze the benefits of the constrained optimization of the acquisition function. Our purpose is to compare empirical performances across (a) different characterizations of the feature spaces, e.g. linear\slash nonlinear subspaces; (b) different properties of the objective function, e.g. additivity\slash non additivity; (c) a real problem set. 

\textbf{Approaches.} We compare our approach (MGPC-BO) with the random embeddings optimization (REMBO)~\cite{Wang2013b}, which performs BO on a random linear subspace of the inputs. Additional baselines include additive models (ADD-BO) \cite{kadasamy_add}, which assumes an additive structure (across dimensions) of the objective $f_{X}$, and one recently proposed VAE-based model (VAE-BO) \cite{GomezBombarelliAutomatic2017} that learns a feature space with deep networks offline. We also include a version of our model presented in Figure \ref{fig:GP-AE} (HMGPC-BO) that uses a hierarchical ICM for the input reconstruction mapping $\mathbf{g}$. The hierarchical ICM partitions the data space into low-dimensional disjoint subsets, i.e. $ \{\mathcal{X}_{i}\}_{i=1}^{Q}$, $\mathcal{X}_{i}\subset\mathbb{R}^{3}$, and assumes independence between reconstructions of different subsets, i.e. $\tilde{\mathbf{x}}^{(i)}\perp \tilde{\mathbf{x}}^{(j)}$, where $\tilde{\mathbf{x}}^{(i)}\in\mathcal{X}_{i}$,
$\tilde{\mathbf{x}}^{(j)}\in\mathcal{X}_{j}$ for $i \neq j$. Moreover, the baselines MGP-BO and HMGP-BO correspond to same modeling as in MGPC-BO and HMGPC-BO, respectively, but without applying the nonlinear constraint in  (\ref{Lconstraint}). We also compare with a different parametrization of the covariance function of the decoder $\mathbf{g}$. The baseline DMGP-BO and DMGPC-BO define a single kernel $k_{c}$ for the reconstruction task while HMGP-BO and HMGPC-BO define different kernels $\{k_{c}^{i}\}_{i=1}^{Q}$, one for each subset of the partitioning. Here, DMGPC-BO and DMGP-BO denote the baseline with and without Lipschitz regularization, respectively. For all the approaches we specify the dimensionality $d_{fs}$ of a feature space where the optimization is performed. Note that this value may differ from the intrinsic dimensionality $d$ of the objective functions i.e. $d_{fs}\neq d$.

\begin{figure*}
    \centering
    \subfigure{\includegraphics[width=1.\linewidth]{Legend_for_all.pdf}}
    \setcounter{subfigure}{0}
    \subfigure[EI]{\includegraphics[width=.3\linewidth]{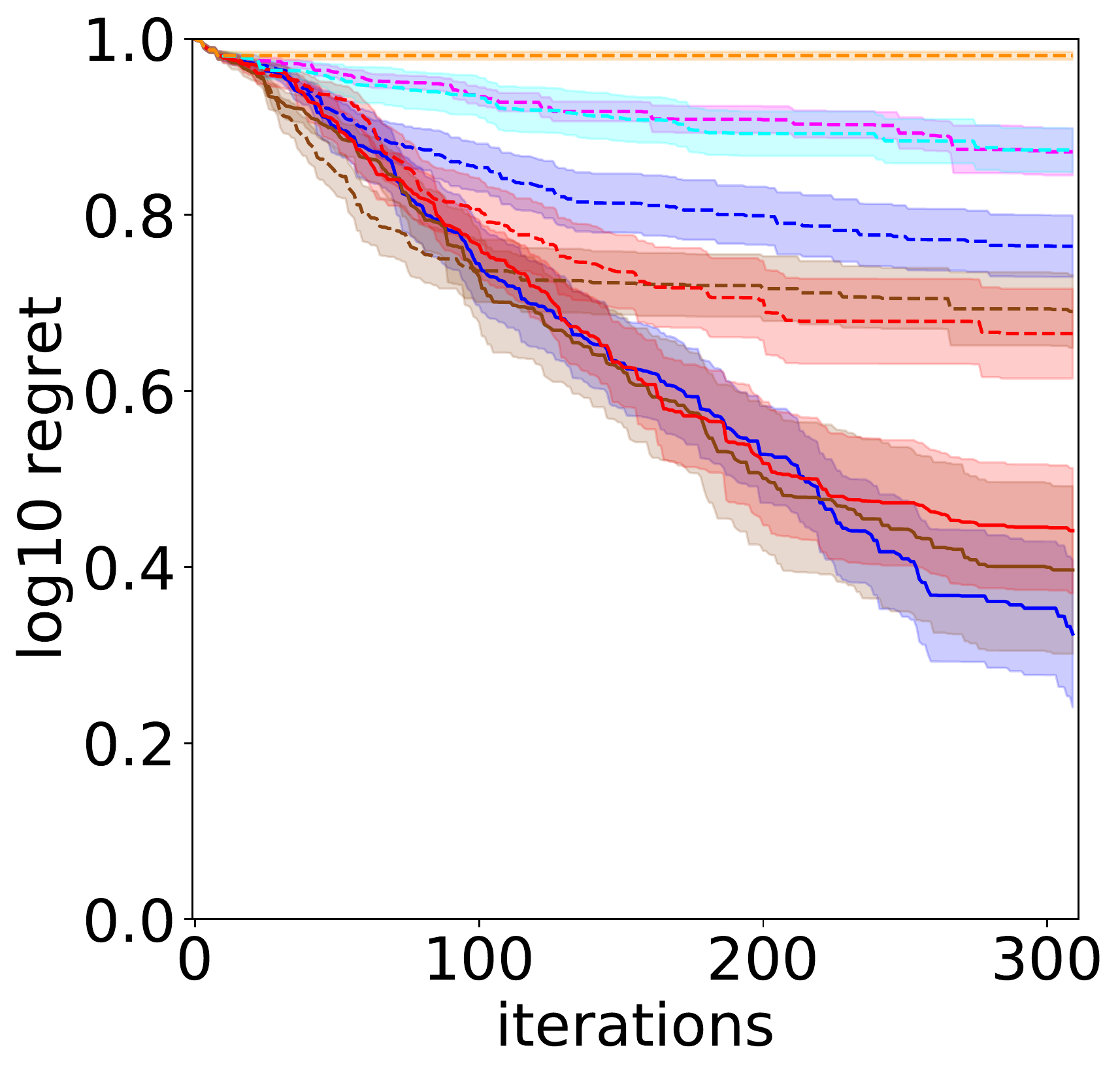}\label{fig:PS_Linear_EI}}%
    \hfill
    \subfigure[UCB]{\includegraphics[width=.3\linewidth]{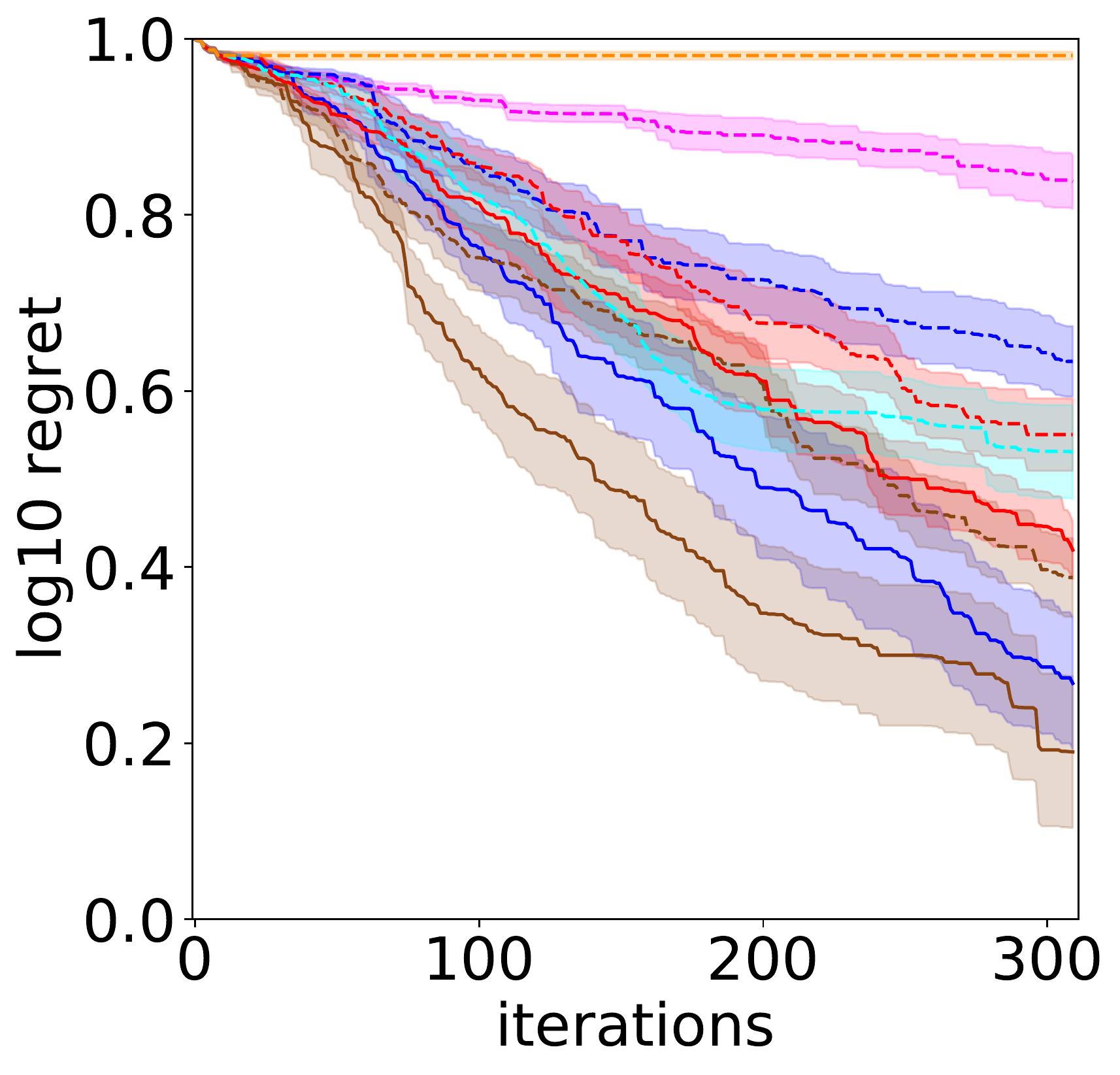}\label{fig:PS_Linear_UCB}}%
    \hfill
    \subfigure[PI]{\includegraphics[width=.3\linewidth]{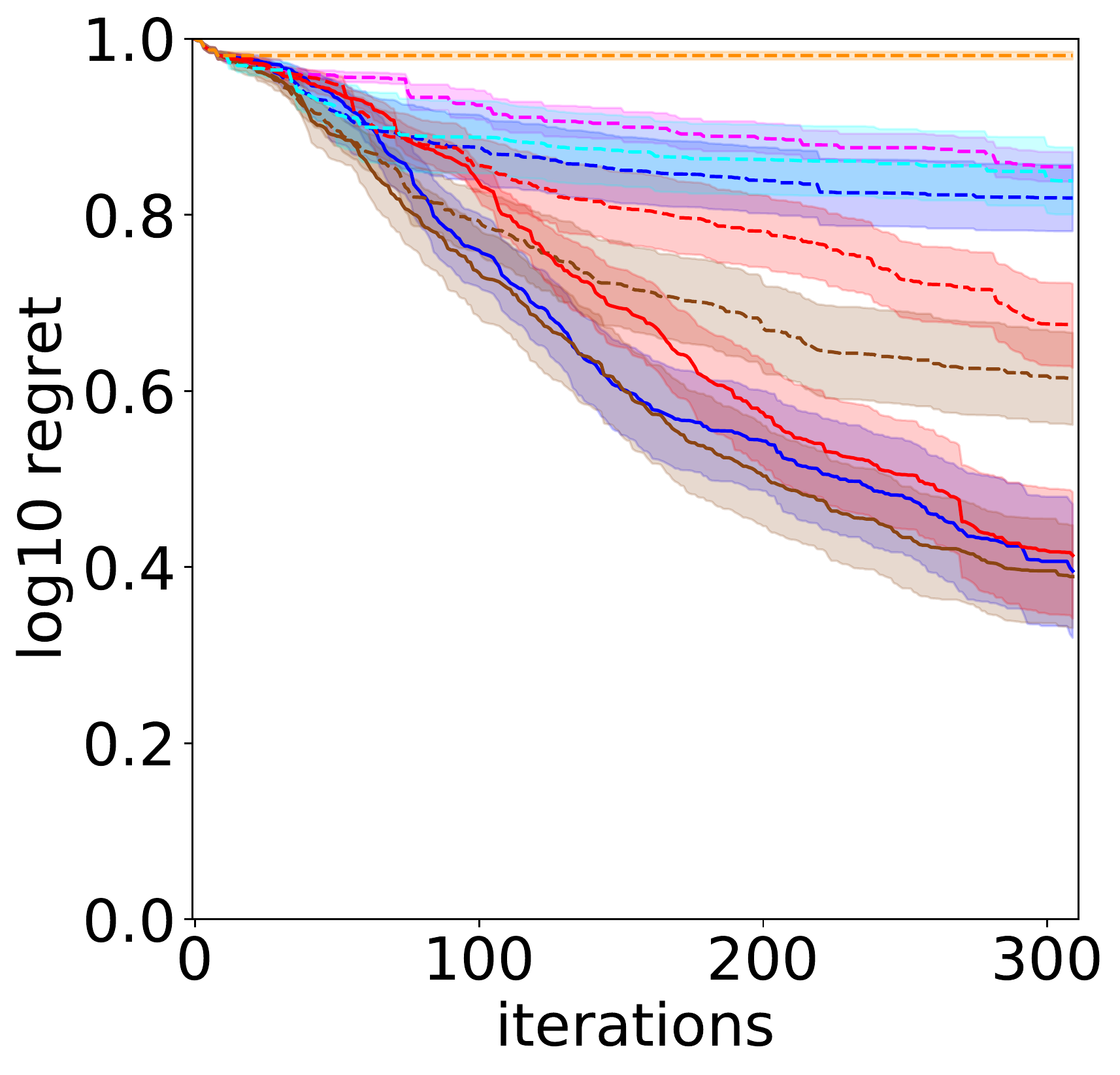}\label{fig:PS_Linear_PI}}
    
    \caption{Optimization progression on product of sines characterized by linear embedding with EI \ref{fig:PS_Linear_EI}, UCB \ref{fig:PS_Linear_UCB} and PI \ref{fig:PS_Linear_PI}. Baselines MGPC-BO, HMGPC-BO and DMGPC-BO learn low-dimensional representations of the objective that are useful for optimization.}\label{fig:PS_Linear} 
\end{figure*}
\textbf{Acquisition functions.} We evaluate the performances of all baselines across common acquisition functions: EI \cite{Mockus1975OnExtremum}, UCB \cite{Srinivas2010} and PI \cite{Kushner1964}, which are also given  in~\eqref{equationPI}--\eqref{equationUCB}. The motivation in selecting the above acquisition functions is that we wish to explore performances of our BO approach on a range of different decision strategies: aggressive exploitation (PI), aggressive exploration (UCB) and one-time-step optimal selection (EI). 
In our experiments we set the $\beta_{t}$ parameter of UCB in (\ref{equationUCB}) to $\sqrt{3}$. Moreover, we do not have access to the true $f_{\text{min}}$ required in  (\ref{equationPI}) and (\ref{equationEI}). Therefore, we compute the improvement based acquisitions (EI and PI) using $y_{\text{min}}:=\min \vec y_{t}$, which is the best noisy observation obtained up to iteration $t$.
The maximization of the acquisition function is identical for all baselines: we first perform a random search step with $5,000$ samples drawn uniformly at random and select the best $100$ locations to apply gradient-based optimization from these starting locations. For box-constrained acquisition optimization we use L-BFGS-B  \cite{Lu1994,Zhu1997AlgorithmOptimization}. For constrained acquisition optimization with nonlinear constraints we use a trust-region interior point method \cite{trust_regionOPT1999}.

\textbf{Model parameters.} In our experiments, we select the \emph{Mat\'ern}$_{5/2}$ kernel as the covariance function for the GPs in each baseline. For the neural network employed in the encoder, the architecture was a single hidden layer with $20$ units, and as the activation function we use the sigmoid activation $1/(1+\exp(-x))$.

\textbf{Experiment setup.} Each BO progression curve shows the mean and standard error of the immediate logarithmic regret $\log_{10}|f(\mathbf{x}_{\text{best}}(t))-f_{\text{min}}|$, where $f_{\text{min}}$ is the true minimum of $f_{X}$ and $\mathbf{x}_{\text{best}}(t)\in\arg\min_{i=1:t} f_{X}(\mathbf{x}_{i})$. Mean and standard error are computed over $20$ experiments with different random initializations. All optimization experiments start with a budget of $10$ data points and perform a total of $300$ iterations. The noise variance is $\sigma^{2}_{n}=10^{-4}$.  



\subsection{Linear feature space}
\label{sec:LinearFeatureSpace}

We consider benchmark functions that are defined in a $d=10$-dimensional space. We map their input space to a $D=60$-dimensional space using an orthogonal matrix $\vec R^{d\times D}$ so that the overall objective is $f_{X}(\mathbf{x})=f(\vec z) = f(\mathbf{R}\mathbf{x})$. 

\subsubsection{Additive objective}
\label{sec:add}

We minimize the \emph{Rosenbrock} benchmark function 
\begin{align} f(\mathbf{z}) = \sum\nolimits_{i=1}^{d-1}[100(z_{i+1}-z_{i}^{2})^{2} + (z_{i}-1)^{2}]
\end{align}
in a $d_{fs}=10$-dimensional feature space. Figure~\ref{fig:RosLinear} shows that HMGPC-BO baseline descends quickly to relatively low regret in the early stages of optimization and recovers better regret at termination than the unconstrained baseline HMGP-BO. 
The VAE-BO baseline improves quickly but lacks exploration due to an insufficiently expressive reconstruction mapping from feature space to data space. We highlight that the VAE-BO model was trained on a budget of $500$ inputs-observations pairs prior to starting the BO experiments. This additional budget, however, still does not allow the VAE-BO to compare well with baselines that learn a feature mapping during optimization. REMBO shows a competitive descent for two main reasons: the fact that the baseline conforms with the linear embedding assumption that characterizes the objective function and the employment of an orthonormal linear mapping which is supposed to improve performances and conforms to structural assumption about the linear embedding $\vec R$. The ADD-BO baseline suffers from the coupling effects of the linear dimensionality reduction $\mathbf{R}$. Overall, Figure~\ref{fig:RosLinear} highlights the fast learning of feature space representations that are effective for optimization with MGPC-BO, HMGPC-BO and DMGPC-BO baselines.

\begin{figure*}
    \centering
    \subfigure{\includegraphics[width=1.\linewidth]{Legend_for_all.pdf}}
    \setcounter{subfigure}{0}
    \subfigure[EI]{\includegraphics[width=.3\linewidth]{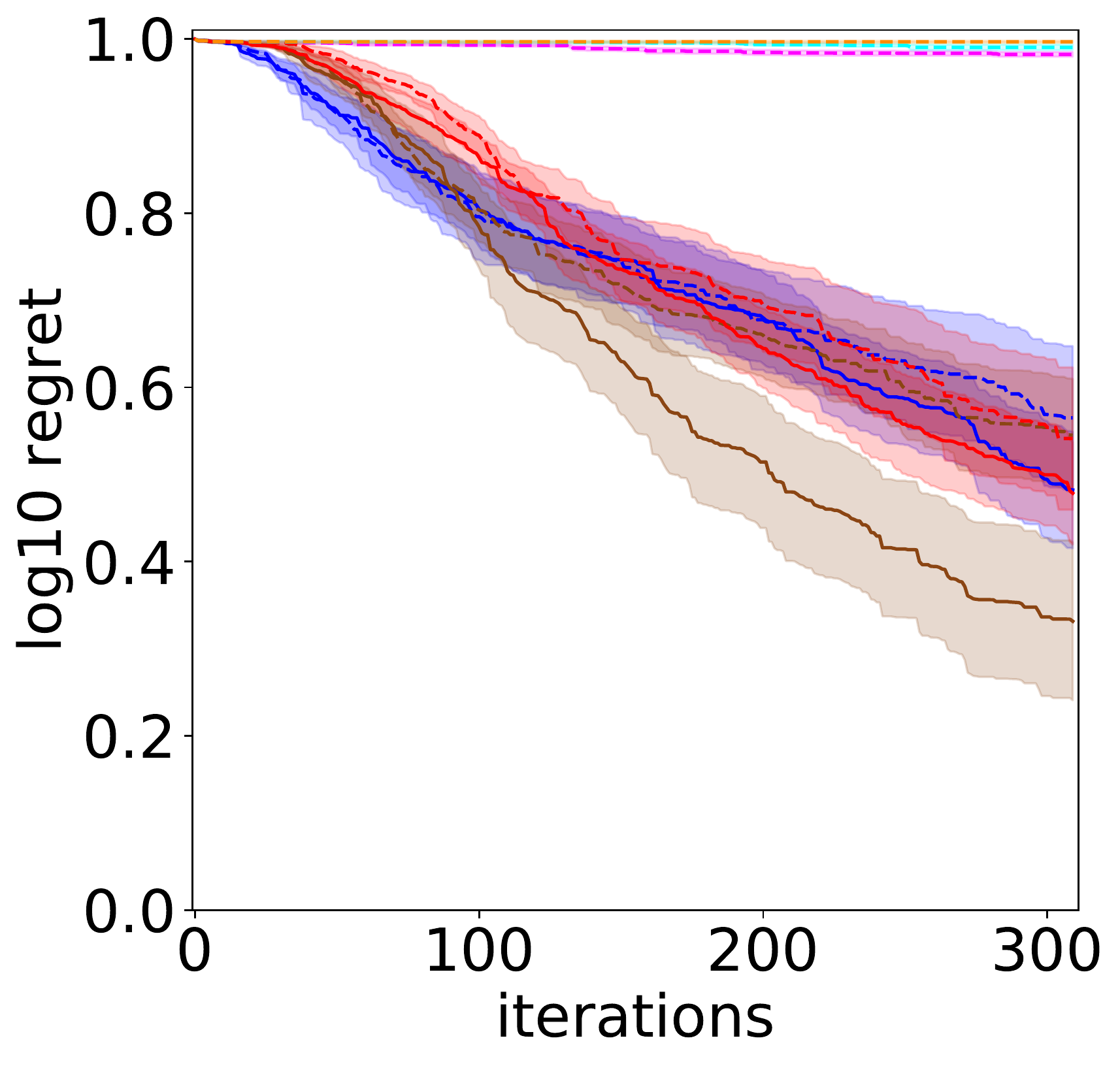}\label{fig:PS_NL_embedding_EI}}%
    \hfill
    \subfigure[UCB]{\includegraphics[width=.3\linewidth]{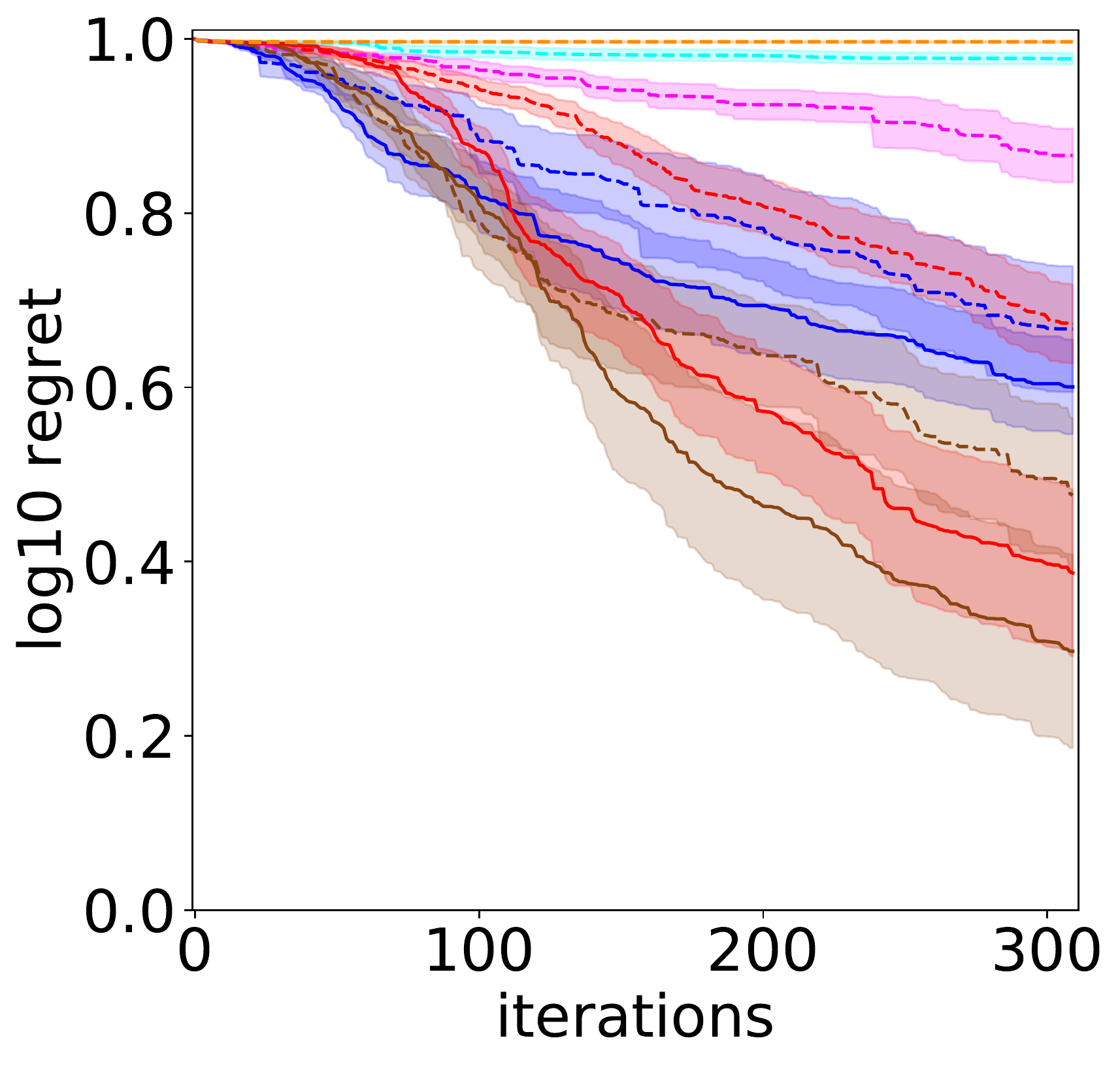}\label{fig:PS_NL_embedding_UCB}}%
    \hfill
    \subfigure[PI]{\includegraphics[width=.3\linewidth]{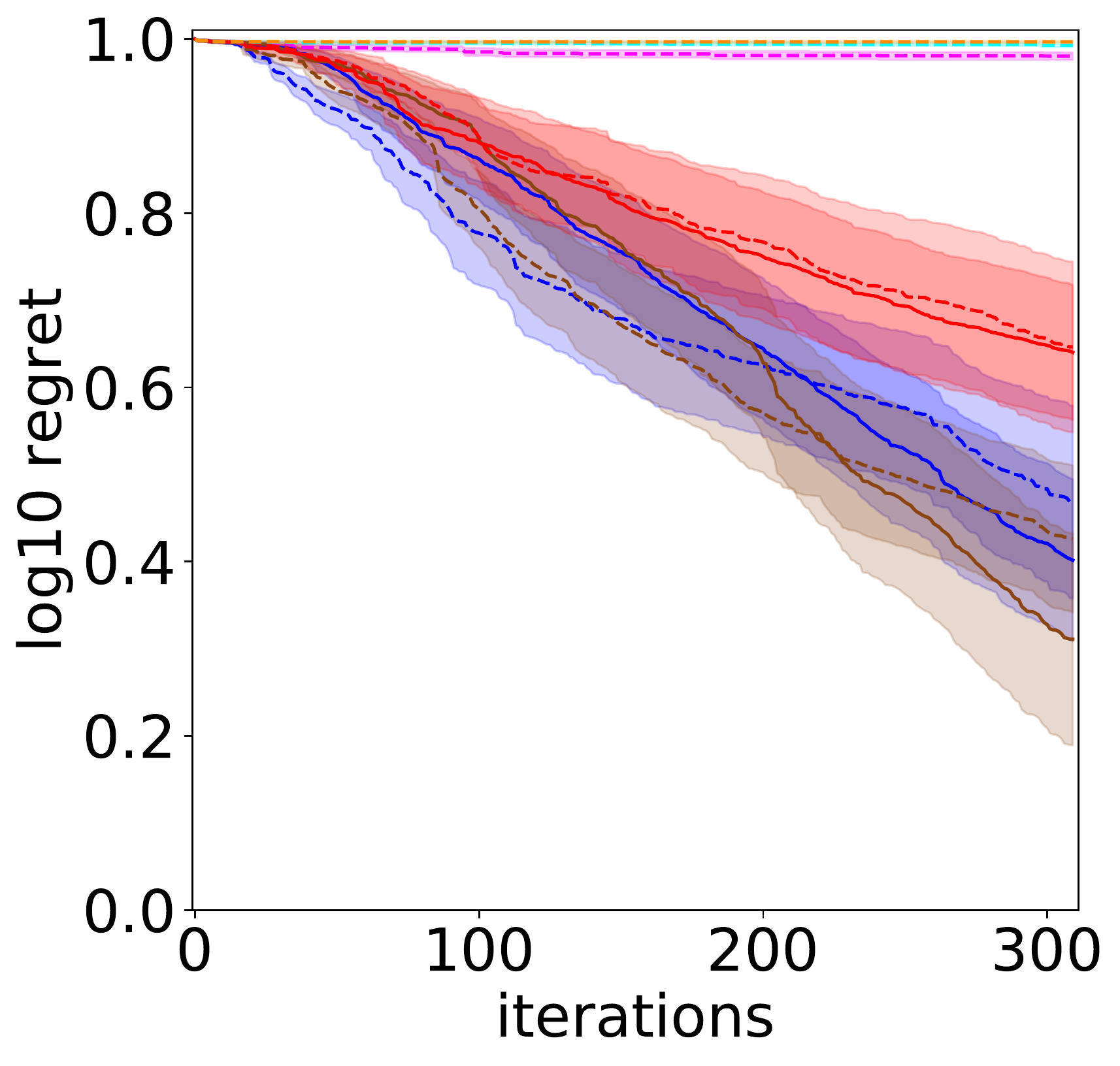}\label{fig:PS_NL_embedding_PI}}
    
    \caption{BO performances expressed as log regret of the product of sines function in a nonlinear embedding. Results are shown for EI \ref{fig:PS_NL_embedding_EI}, UCB \ref{fig:PS_NL_embedding_UCB} and PI \ref{fig:PS_NL_embedding_PI}.
    All our baselines with nonlinear constraint, namely MGPC-BO, DMGPC-BO and HMGPC-BO learn useful representations in feature space for optimization. There is highly significant difference of $0.014\%$ between DMGPC-BO and ADD-BO. }\label{fig:PS_NL_embedding}
\end{figure*}



\subsubsection{Non-additive objective}
\label{sec:non_add}

Here, we optimize the \emph{product of sines} with intrinsic dimensionality $d=10$ 
\begin{align}
f(\mathbf{z}) = 10\sin(z_{1}) \prod\nolimits_{i=1}^{d}\sin(z_{i})
\end{align}
and compare results when the additivity assumption is not satisfied. Figure~\ref{fig:PS_Linear} shows the regret curves obtained optimizing the objective on a $d_{fs}=10$-dimensional feature space. Solid lines describe the Lipschitz-regularized baselines MGPC-BO, HMGPC-BO and DMGPC-BO (with nonlinear constraint), while dashed lines are baselines that apply box-constrained maximization of the acquisition in feature space. The HMGP-BO, MGP-BO and DMGP-BO regrets flatten early in both improvement-based acquisition functions (EI and PI) since these acquisition functions highlight locations in feature space that are too far away from the training data. In this setting, the decoder $\mathbf{g}$ returns the same high-dimensional reconstruction, which prevents BO from exploring. The constrained maximization of the acquisition is beneficial for all our models. We also note that the REMBO baseline conforms to the intrinsic linear low-dimensionality assumption described in Section \ref{sec:LinearFeatureSpace} and is  the most competitive baseline especially for UCB acquisition. However, the linear reconstruction mapping applied by REMBO also suffers from non-injectivity, and this slows down exploration in the high-dimensional space. The linear projection  deteriorates performances of the additive model. ADD-BO assumes independence between axis-aligned projections of the high-dimensional space, while the linear mapping $\mathbf{R}$ couples all subsets of dimensions. This linear mapping, therefore, penalizes optimization with independent additive components. The VAE-BO approach requires much larger amounts of data to learn a meaningful reconstruction mapping than available in our experiment. Thus, most locations in feature space are mapped to similar reconstructions. This explains the flat curve observed on all VAE-BO progressions with different acquisitions. 

\subsection{Nonlinear feature space with non-additive objective}
\label{sec:NonLinearFeatureSpace}
We consider the product of sines functions and apply a nonlinear dimensionality reduction. We define a single-layer neural network mapping to elevate the dimensionality of the objective to $D=60$, i.e. $f_{X}(\mathbf{x})=f(\gamma(\mathbf{R}\mathbf{x}))$. Here $\gamma$ is the sigmoid activation function. We select a dimensionality of the feature space as in previous sections $d_{fs}=10$ which is equal to the intrinsic dimensionality of the objective function $d=10$. Figure \ref{fig:PS_NL_embedding} shows the progression of the regret over 300 BO iterations. We can observe consistent improvements of MGPC-BO, HMGPC-BO and DMGPC-BO with respect to VAE-BO which also assumes a nonlinear embedding for the objective.  The performance of MGPC-BO, HMGPC-BO and DMGPC-BO also retain better regret at termination than with box-constrained acquisition maximization (MGP-BO, HMGP-BO, DMGP-BO). Here we apply a significance testing with the Wilcoxon signed-rank test \cite{wilcoxon1992individual} at termination of the optimization between the best performing of our baselines, namely DMGPC-BO and the best competitive baseline that is ADD-BO. We observe a significance of at least $0.014\%$ for all acquisition functions (largest p-value $p=0.00014$ for UCB acquisition) meaning that our best baseline DMGPC-BO is highly significantly different than the ADD-BO baseline and attains better regret than ADD-BO at termination of the optimization. 

Overall, we observe that the constrained maximization of the acquisition function is beneficial for the proposed model. The advantages with respect to ADD-BO, REMBO and VAE-BO baselines are more evident with the product of sines objective with nonlinear embedding while with the Rosenbrock we retain no worse regret.



\subsection{Sensitivity analysis on real data}

\begin{figure*}
    \centering
    \subfigure{\includegraphics[width=1\linewidth]{Legend_for_all}}
    \setcounter{subfigure}{0}
    \subfigure[$d_{fs}=2$, PI]{\includegraphics[width=.24\linewidth]{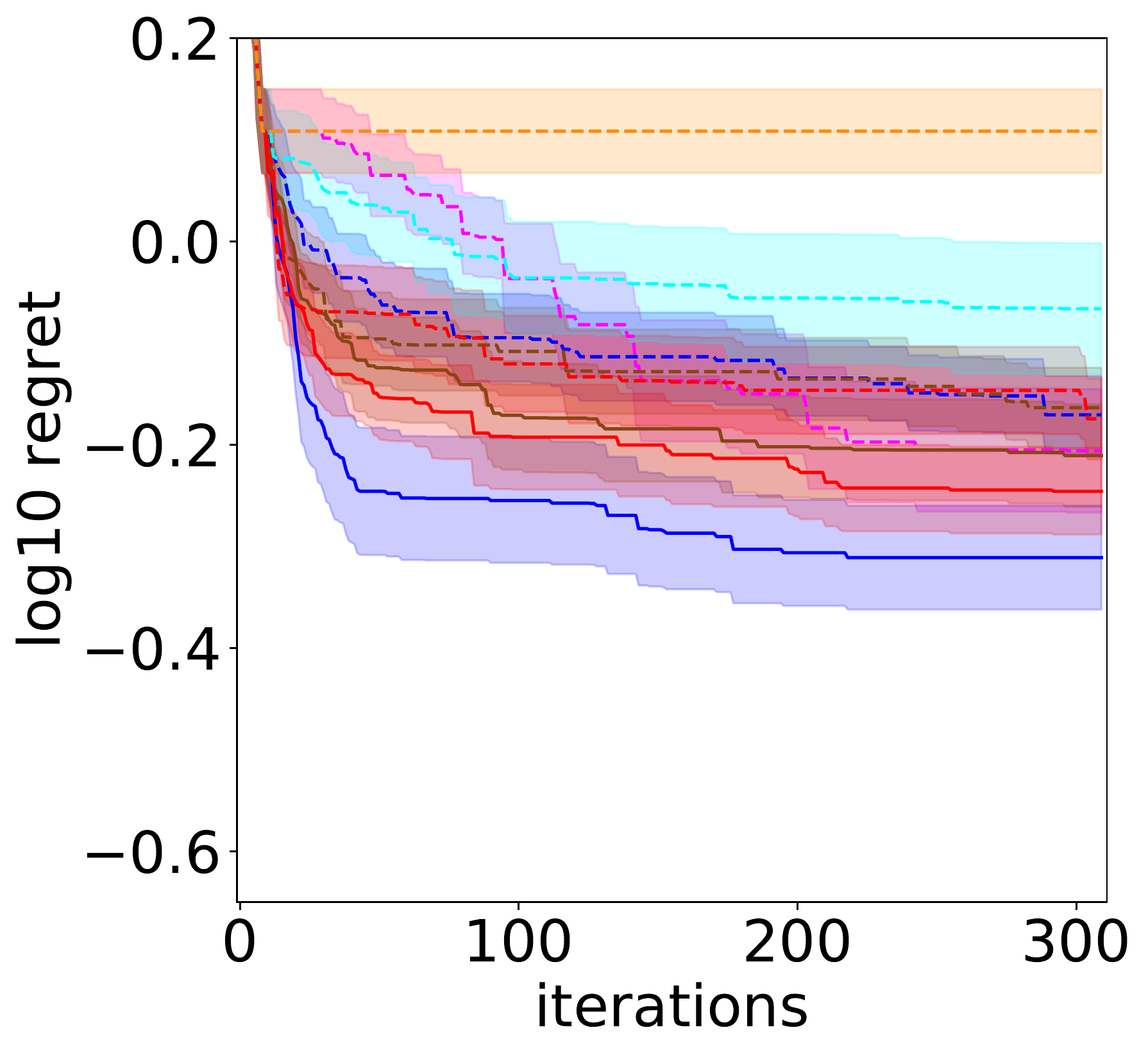}}%
    \hfill
    \subfigure[$d_{fs}=3$, PI]{\includegraphics[width=.24\linewidth]{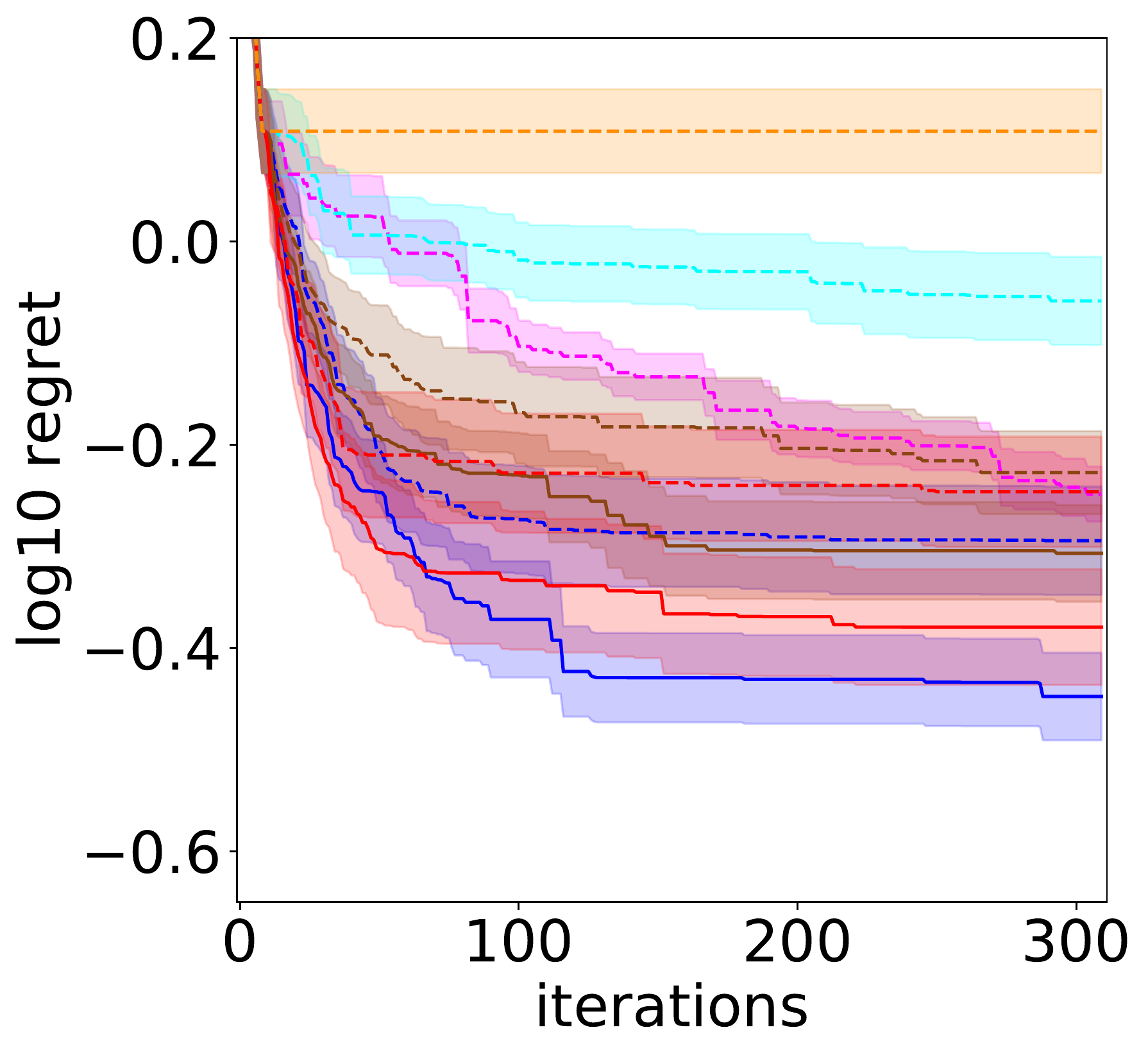}}%
    \hfill
    \subfigure[$d_{fs}=4$, PI]{\includegraphics[width=.24\linewidth]{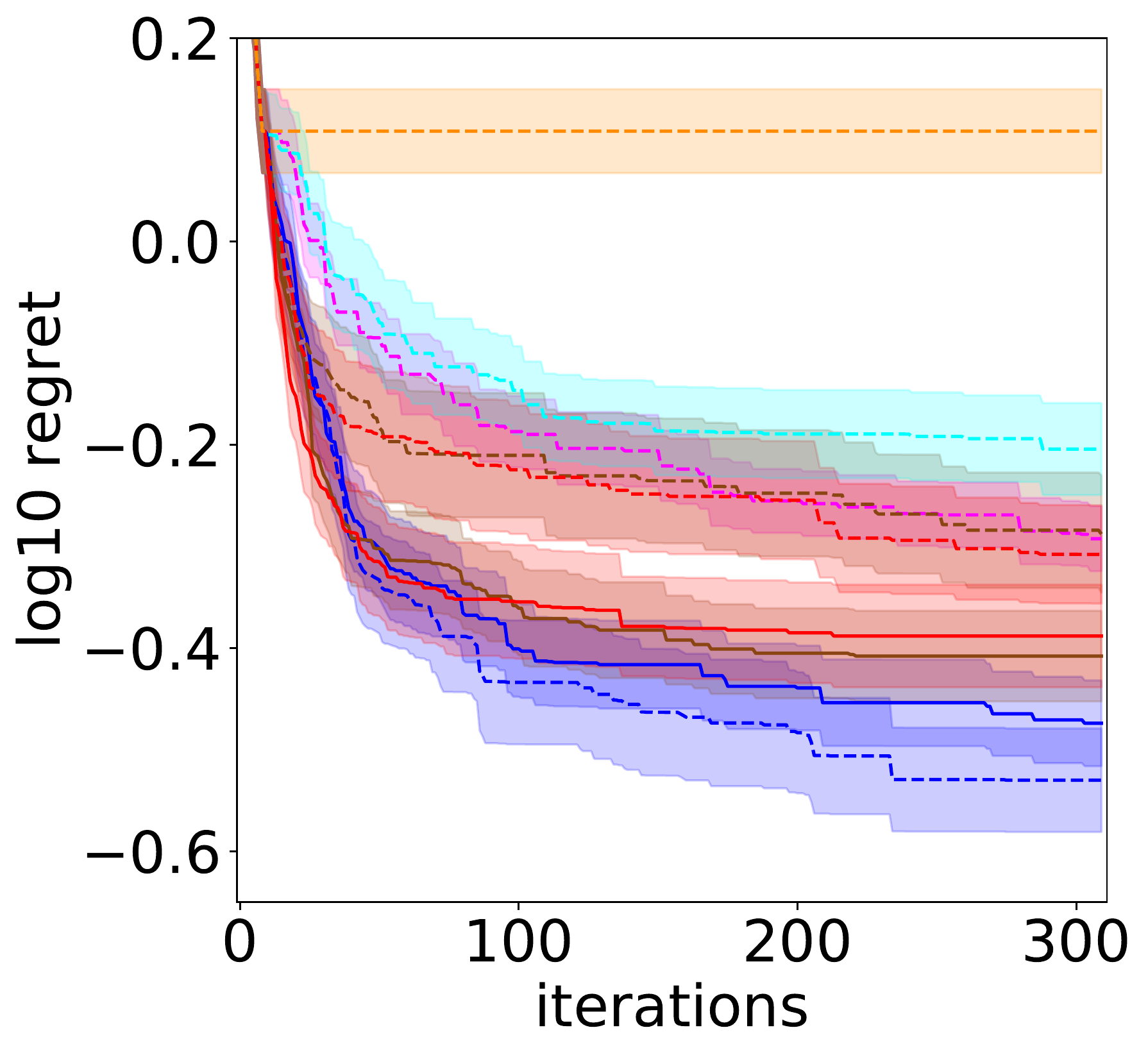}}
    \hfill
    \subfigure[$d_{fs}=6$, PI]{\includegraphics[width=.24\linewidth]{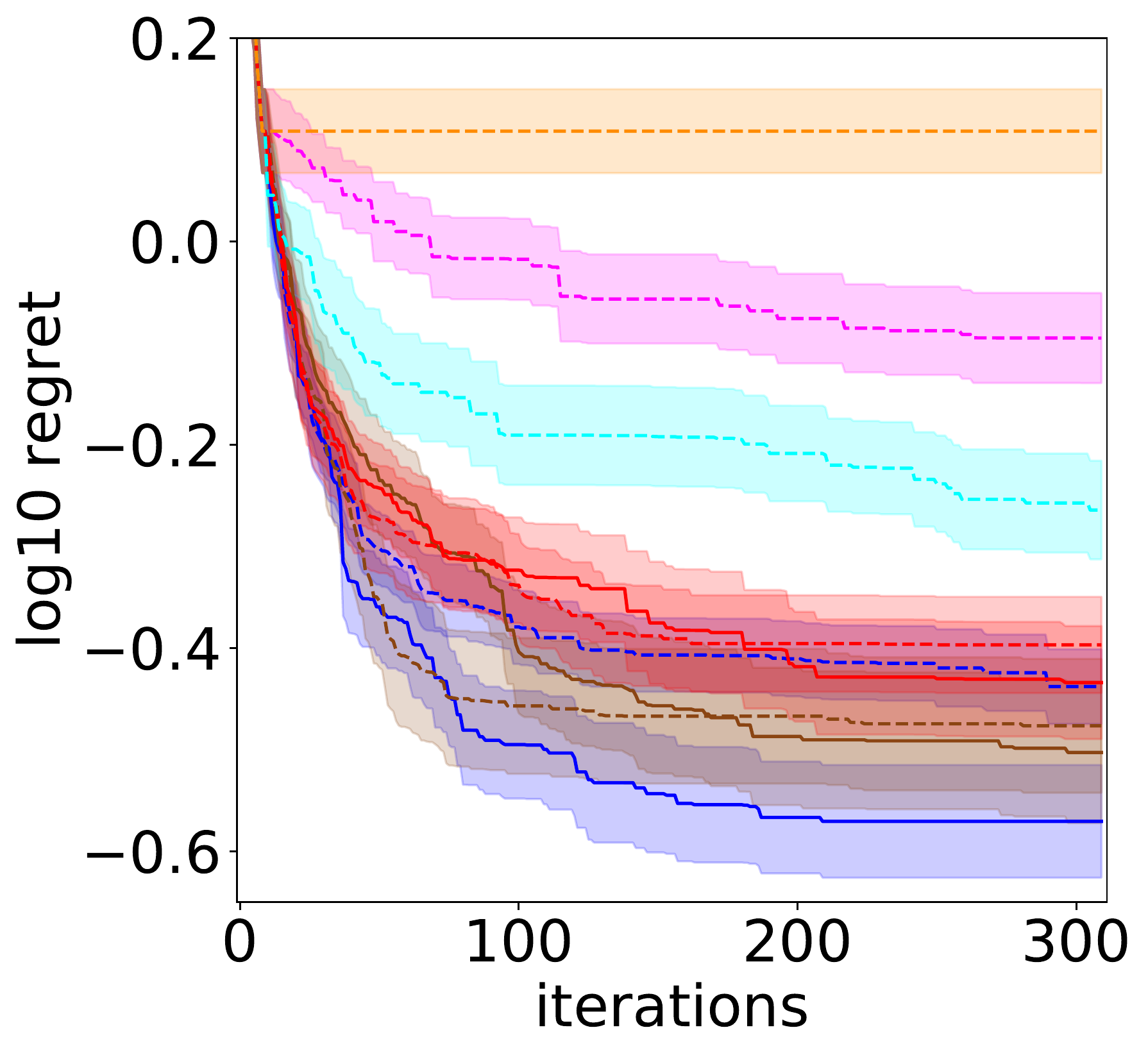}}
    \caption{Sensitivity analysis with respect to the dimensionality of the feature space $d_{fs}$ on a real problem set. We test all approaches on a set of feature space dimensionalities $d_{fs}=2, 3, 4, 6$. The performances of our baselines clearly deteriorate for $d_{fs}=2$. Our baseline MGPC-BO show better performances than the best competing baseline ADD-BO and REMBO and reach the minimum in notably less iterations.}\label{fig:sensitivity}
\end{figure*}

Here we apply a sensitivity analysis with respect to the dimensionality of the feature space $d_{fs}$ on a $D=12$-dimensional real problem. We consider the Thomson problem of finding the lowest potential configuration of a set of electrons on a sphere \cite{dolan2004benchmarking}. This is a central problem in physics and chemistry for identifying a structure with respect to atomic locations \cite{dolan2004benchmarking}. The potential of a set of $n_{p}$ electrons on a unit sphere is given by the objective
\begin{equation}
    \sum\nolimits_{i=1}^{n_{p}-1}\sum\nolimits_{j=i+1}^{n_{p}}((x_{i}-x_{j})^{2}+(y_{i}-y_{j})^{2}+(z_{i}-z_{j})^{2})^{-1/2}.
\end{equation}
This is a constrained minimization problem with constraints $x_{i}^{2}+y_{i}^{2}+z_{i}^{2}=1$ for $i=1,...,n_{p}$, which means that all electrons must lie on a unit sphere. We represent the variables of the problem as spherical coordinates with unit radius. This allows us defining two variables per point with a total number of $2n_{p}$ (azimuthal and polar angles) parameters to optimize within box constraints. For optimization, we select $n_{p}=6$, which results in a $D=12$-dimensional problem and we optimize it on low-dimensional feature spaces of dimensionalities $d_{fs}=6, 4, 3, 2$ to observe the effect of this hyper-parameter in the optimization. Figure \ref{fig:sensitivity} shows a comparison of our approaches with ADD-BO, REMBO and VAE-BO baselines on a single acquisition function PI. Overall, we observe a deterioration of performances with diminishing dimensionality of the feature space. The regret clearly increases for our baselines when we select $d_{fs}=2$ meaning that, with a high compression rate, the probabilistic model for MGPC-BO, DGPC-BO and HMGPC-BO learns less useful features for optimization. We observe the most competitive baseline to be ADD-BO, which decomposes the $12$-dimensional problems into $D/d_{fs}$ sub-problems with dimensionality $d_{fs}$. Another competitive baseline is REMBO. We apply a significance test and compare our nonlinearly constrained baseline MGPC-BO with the most competitive baseline (ADD-BO or REMBO) for each plot of Figure \ref{fig:sensitivity} at termination of the optimization. We select the Wilcoxon signed-rank test \cite{wilcoxon1992individual}, which does not assume that the difference between the sample populations is Gaussian. 
For feature space dimensionality $d_{fs}=2$ we do not observe values significantly different since the $p$-value is $p=0.135$. This is due to the deterioration of performances at $d_{fs}=2$. For $d_{fs}=3$ we observe a more significant difference between MGPC-BO and ADD-BO with p-value $p\leq 0.002$. With hyper-parameter values $d_{fs}\geq 4$ we observe significantly different baselines with significance at $0.6\%$ (difference between MGPC-BO and ADD-BO for $d_{fs}=4$ with $p$-value $p\leq 0.003$ and between MGPC-BO and REMBO for $d_{fs}=6$ with p-value $p\leq 0.006$).
Overall, we observe our constrained baselines to perform better than ADD-BO and REMBO and to reach the lowest value in notably less BO iterations.
\subsection{Run-time complexity}

The computational complexity of MGPC-BO is $\mathcal{O}(D^3+N^3)$ due to the eigen-decomposition of both the coregionalization ($D^3$) and kernel matrix ($N^3$). The baseline HMGPC-BO scales with $\mathcal{O}(d_{out}^3Q+N^3Q)$ with $Q$ being the number of independent subsets of dimensions, i.e. $Q=D/d_{out}$, with $d_{out}$ being a small constant value ($d_{out}=3$). This baseline achieves faster computations when having small number of data points $N$, for large number of data points and large number of dimensions (both tending to infinity) the MGPC-BO results more efficient. The baseline DMGPC-BO instead has complexity $\mathcal{O}(d_{out}^{3}Q+N^3)$, which is  faster than the MGPC-BO. MGP-BO, DMGP-BO and HMGP-BO have the same complexity of MGPC-BO, DMGPC-BO and HMGPC-BO, respectively. The remaining baselines have all computational complexity $\mathcal{O}(N^3)$ due to the matrix inversion of the covariance matrix for GP training which is used in ADD-BO, REMBO and VAE-BO. Our baseline has an additional overhead of at least a linear term $d_{out}^3Q$, which implies slower training times for our probabilistic model. This is a reasonable trade off for improved optimization performances and better data efficiency in our reconstruction model.

\section{Conclusion}

We proposed a framework for efficient Bayesian optimization of intrinsically low-dimensional black-box functions based on nonlinear embeddings. In our model, a manifold GP learns useful low-dimensional feature representations of high-dimensional data by jointly learning the response surface and a reconstruction mapping. 
Our approach allows for optimizing acquisition functions in a low-dimensional feature space. 
Since exploration in feature space (driven by the acquisition function) does not necessarily mean exploration in the high-dimensional parameter space, we introduce a nonlinear constraint based on Lipschitz continuity of predictions of the reconstruction mapping, which encourages exploration in the vicinity of the training data and mitigates un-identifiability issues in data space, which hinder optimization.

\section*{Acknowledgments}
We thank James T. Wilson for valuable feedback on early drafts of the manuscript. This work has been supported by the EPSRC Centre for Doctoral Training in High Performance Embedded and Distributed Systems (HiPEDS, Grant EP/L016796/1) and the Data Science Institute, Imperial College London.

\bibliographystyle{spmpsci}      
\bibliography{references_ECML2020}   

\end{document}